\documentclass[11pt,a4paper]{article}
\usepackage[nohyperref]{acl2018}
\usepackage{times}
\usepackage{latexsym}
\usepackage{amsmath,amssymb}

\aclfinalcopy
  
\usepackage{booktabs,tabularx,rotating}
\usepackage{graphics,subfigure,graphicx}
\usepackage{tikz,pgflibraryshapes}
\usepackage{lscape} 

\usepackage{url}

\newcommand{\Ni}{({\em i})~}
\newcommand{\Nii}{({\em ii})~}
\newcommand{\Niii}{({\em iii})~}
\newcommand{\Niv}{({\em iv})~}

\newcommand{\wt}{{\tt WikiTailor~}}
\newcommand{\wtend}{{\tt WikiTailor}}
\newcommand{\wtV}[1]{50-WT#1}
\newcommand{\wtVI}[1]{60-WT#1}
\newcommand{\wtVall}{50-WT*}
\newcommand{\wtVIall}{60-WT*}
\newcommand{\wtall}[1]{*-WT#1}
\newcommand{\irV}[1]{50-IR#1}
\newcommand{\irC}[1]{100-IR#1}
\newcommand{\irVall}{50-IR*}
\newcommand{\irCall}{100-IR*}
\newcommand{\irall}[1]{*-IR#1}

\newcommand{\clesa}{$d_{\rm ESA}$}
 
\newcommand{\mc}[3]{\multicolumn{#1}{#2}{#3}}

\definecolor{blueplot}{HTML}{6a3d9a}
\definecolor{redplot}{HTML}{e31a1c}
\definecolor{orangeplot}{HTML}{ff7f00}
\definecolor{greenplot}{HTML}{33a02c}

\title{Tailoring and Evaluating the Wikipedia for \\in-Domain Comparable Corpora Extraction}

\author{Cristina Espa\~na-Bonet$^1$, Alberto Barr\'on-Cede\~no$^2$ and Llu\'{i}s M\`arquez$^3$\thanks{~ Work conducted while this author was at the Qatar Computing Research Institute (QCRI-HBKU).} \\
$^1$ DFKI GmbH, Saarbr\"uken\\
$^2$ Universit\`a di Bologna, Forl\`i, Italy\\
$^3$ Amazon, Barcelona  
}

\begin{document}
\maketitle

\begin{abstract}%
We propose an automatic language-independent graph-based method to build 
\textit{\`a-la-carte} article collections on user-defined domains from the Wikipedia. The core model is based on the exploration of the encyclopaedia's category graph and can produce both monolingual and multilingual comparable collections.
We run thorough experiments to assess the quality of the obtained corpora in 10 languages and 743 domains. 
According to an extensive manual evaluation, our graph-based model outperforms a retrieval-based approach and reaches an average precision of $84\%$ on in-domain articles. 
As manual evaluations are costly, we introduce the concept of \emph{domainness} and design several automatic metrics to account for the quality of the collections.
Our best metric for domainness shows a strong correlation with the human-judged  precision, 
representing a reasonable automatic alternative to assess the quality of domain-specific corpora.
We release the \wt toolkit with the implementation of the extraction methods, the evaluation measures and several utilities. \wt  makes obtaining multilingual in-domain data from the Wikipedia easy. 
\end{abstract}

\section{Introduction}
\label{sec:intro}

Different natural language processing (NLP) and information retrieval (IR) tasks require large amounts of domain-specific text with different levels of parallelism. With such data, one can obtain in-domain lexicons, semantic representations of concepts, train specialised machine translation engines or question answering systems.
A common strategy to gather multilingual domain-specific material is crawling the Web; 
e.g., looking for different language editions of a website~\cite{Resnik:03,Espla:09}.
Nowadays, one of the largest controlled sources for this kind of text at the fingertips is the Wikipedia 
---an online encyclopaedia with millions of topic-aligned articles in multiple languages.%
\footnote{\url{http://www.wikipedia.org}} 
Wikipedia is not only comparable, but some fragments, and even full articles, are parallel across languages due to cross-language (CL) text re-use.%
\footnote{An interesting discussion about the value of the Wikipedia as a comparable corpus was carried out through the Corpora list \url{https://mailman.uib.no/public/corpora/2014-June/020621.html}; last visited: May 2020.}


In this paper, we explore the value of the Wikipedia as a source for domain-specific comparable text with a practical perspective. 
We present a methodology to extract in-domain articles by taking advantage of Wikipedia's categorisation mark-up and its graph structure.
The multilingual aspect of the resource facilitates the extraction of multilingual counterparts. 

In our experiments, we extract collections with different systems in 10 languages and 743 domains, and manually evaluate the adequacy to the domain for a subset of the collections. 
%
Nevertheless, manual evaluations are expensive both in terms of time and money. An automatic evaluation is problematic in this area since, to our knowledge, there is no accepted way to measure how well a collection represents a domain. To this end, we define the concept of \emph{domainness} as a combination of the representativity and cohesion of texts.
We introduce several automatic metrics that model the occurrence, co-occurrence, distribution of the characteristic vocabulary, and the semantic similarity among the articles.

We release the implementation of our architectures and the quality metrics within \wtend, a Java toolkit designed to extract and analyse corpora from Wikipedia in any language and domain. Both a stand-alone executable and the source code are available.%
\footnote{\wt is available at \url{http://cristinae.github.io/WikiTailor/}}
As a result, the generation of comparable resources becomes relatively easy.
We also make available some of the collections generated in our analysis. We share the domain-specific term vocabularies and the identifiers of the articles obtained with our best model for all the domains in the languages under study ---English, French, Spanish, German, Arabic, Romanian, Catalan, Basque, Greek, and Occitan.%
\footnote{\url{http://cristinae.github.io/WikiTailor/experiments.html}}
Notice that, contrary to domain names, vocabularies are not parallel but can be useful for other cross-language or multilingual applications. 



The rest of the paper is distributed as follows.
Section~\ref{sec:WP} overviews comparable corpora acquisition methods, with special focus on the categorisation and multilinguality of the Wikipedia. The relevance of Wikipedia for NLP and IR is also highlighted.
Section~\ref{sec:related} summarises some related work and points out similarities and differences with our study.
Section~\ref{sec:models} presents our models for the automatic extraction of (multilingual) in-domain corpora.
Section~\ref{sec:analysis} describes the experimental setting, analyses the characteristics of the collections extracted, and reports the results of our manual evaluation to assess their quality.
In Section~\ref{sec:evaluation}, we define the concept of domainness, introduce several automatic evaluation metrics, and in Section~\ref{sec:automatic} we use them to quantify the quality of the produced collections, and analyse the correlation with human judgements.
We summarise the work and draw our conclusions in Section~\ref{sec:conclusions}.
We include a glossary with Wikipedia-specific terms in Appendix~\ref{app:concepts} and detail the  crowdsourcing experiment that leads to our manual evaluation in Appendix~\ref{app:crowdflower}.


\section{Comparable Corpora and the Wikipedia}
\label{sec:WP}

Multiple kinds of Web contents have been used as a source for the acquisition of comparable corpora. 
Usually, the first stage consists of acquiring the documents on the required languages
\cite{Resnik:03,Talvensaari2008,Aker12alight,Plamada:13}.
The second stage is usually alignment; i.e.\ identifying pairs of comparable documents
\cite{Pouliquen:03,Tao:05,Munteanu:05,Vu:2009:FMD:1609067.1609161,GamalloGonzalez:10}.
Among them, \citet{Plamada:13} and \citet{GamalloGonzalez:10} are specially relevant to this work since they use Wikipedia as a corpus. In this case and up to the limitations we discuss later, alignment is close to trivial due to the existing links between articles in different languages.

In general, three properties cause the Wikipedia particularly suitable as a source of comparable and parallel data:
\Ni it contains editions in a large number of languages,%
\footnote{299 active languages in May 2020.}  
\Nii articles covering the same topics in different language editions are connected via \textit{inter-language links} also called \textit{langlinks},
and \Niii articles have categories which purpose is both describing the topic covered and grouping together related articles.

Nevertheless, it also presents drawbacks. 
First, the inter-language links (as many other characteristics) are subject to inconsistencies because most often they are manually created by volunteers. A volunteer may make mistakes linking non-equivalent concepts; there are even cases in which an article in one edition is linked from two or more articles in another one~\cite{HechtDarreb:10}. 
Second, an article can belong to multiple categories and it is even possible to construct loops with categories; i.e.~no strict tree hierarchy is in place~\cite{zeschGurevych:07}.
Given that categories are built collaboratively, they are arbitrary at times,
many articles are not associated to the categories they should belong to objectively, and one can observe the phenomenon of over-categorization%
\footnote{This phenomenon is stressed in the Wikipedia itself: 
\url{http://en.wikipedia.org/wiki/Wikipedia:Overcategorization}; last visited: May 2020.}.
Consequently, the Wikipedia category graph (WCG) and the links between languages must be considered carefully in order to extract topic-aligned articles across multiple language editions.

Moreover, the intersection across languages tends to be relatively small. In general, smaller Wikipedia editions are not subsets of the larger ones. In the dumps considered for this study, only $0.4\%$ of the articles are common across all 10 editions, which are within the top-100 according to their size.
Among the largest four editions, which represent relatively close cultures (English, French, Spanish, German), the number only grows to $4.8\%$.
\citet{HechtDarreb:10} called this effect \emph{context diversity}. 
According to their analysis, the articles in the intersection correspond to ``globally relevant concepts'', whereas the singletons show cultural diversity.
Recent studies show that the level of diversity across languages remains ---both for text and for images~\cite{He:18}. Consequently, the importance and presence of different topics depends on the language. So, one should expect to be able to obtain more easily comparable corpora for topics associated to the globally relevant concepts.

Even with all this \emph{noise}, which must be acknowledged and taken into account, 
the Wikipedia has been widely and successfully used in (CL)-NLP and (CL)-IR.
For example, it has been used for terminology and bilingual dictionary extraction \cite{Erdmannetal:08,YuTsujii:09,ProchassonFung:11,Chuetal:14,Jakubina:16}. In most of these models, Wikipedia's inter-language links are crucial to obtain an aligned comparable corpus.

The value of the Wikipedia as a source of highly comparable and parallel sentences was soon observed too~\cite{Adafre:06,Yasuda:08,Smithetal:10,Plamada:12,Stefanescuetal:12,Skadinaetal:12,Barronetal:15}. With the rise of 
deep learning for NLP and the need of large amounts of \emph{clean} data, the use of Wikipedia has grown exponentially not only for parallel sentence extraction and machine translation \cite{adamThesis,rameshPrasad:18,ruiterEtAl:2019,schwenkEtAl:2019}, but also for training models to obtain semantic representations of words and sentences.

Word and contextual embeddings have been trained on it, so that the resources are nowadays at hand for more than 100 languages. Examples include fastText \cite{Bojanowski2017,graveEtAl:2018} and MUSE word embeddings~\cite{lampleEtAl:2018}, BERT multilingual embeddings \cite{devlinEtal:2019} and LASER sentence embeddings~\cite{artetxeSchwenk:2019}.

Semantic representations can also be obtained via explicit semantic analysis (ESA)~\cite{Gabrilovich:07} and have been widely used in IR to compute the semantic relatedness of concept vectors. CL-ESA~\cite{HassanMihalcea:2009,Potthast:2008a} is a cross-language extension of the explicit semantic analysis model which allows for computing this semantic relatedness across languages. Compared to neural network-based embeddings, CL-ESA representations are less sensitive to the amount of training data and differences in sizes among languages (see Section~\ref{sub:metrics}) and therefore they are adequate within the multilingual setting we present in this work.

\section{Related Work}
\label{sec:related}

\citet{GamalloGonzalez:10} proposed the basis to extract different kinds of comparable material from the Wikipedia by exploiting its metadata (category tags) and the WCG\@. They distinguished among collections built with: 
\Ni non-aligned articles defined as those which belong to the same topic just because they have associated the same category;
\Nii strongly-aligned articles as those which are connected by an inter-language link and both belong to the same category; and
\Niii softly-aligned articles as those which are connected by an inter-language link but do not necessarily share the same category. 
These extractions are implemented in CorpusPedia\footnote{\url{http://gramatica.usc.es/pln/tools/CorpusPedia.html}}. The tool is designed to extract comparable corpora from the Wikipedia by considering a pair of languages and a category. Given a category, the tool generates the three kinds of comparable corpora by considering every article belonging to it and its sub-categories at one-level depth.

Our work has lots of similarities with theirs. We also consider the Wikipedia as the corpus and exploit its metadata. For the alignment, we opt for the first type, we retrieve all the articles ---even if they are not linked--- in two or more languages which belong to the same domain, extending the method to deal with complete domains instead of with individual categories. In order to extract the domains, we explore WCG and, as a result of avoiding their ``strict'' strategy based on the exact category, we are able to extract more articles. 
This idea was first sketched in~\citet{Barronetal:15}, where we also extracted parallel sentences from the comparable corpora in \textit{Computer science}, \textit{Science} and \textit{Sport} to successfully domain-adapt a machine 
translation system.

The WCG is close to a taxonomy structure~\cite{zeschGurevych:07} and can be easily explored, but the exploration might be slow given the size of some Wikipedia editions, the high density of the graph, and the existence of loops. Several works facilitate the task. 
PetScan\footnote{\url{https://petscan.wmflabs.org/}} is an online utility that retrieves all the articles that depart from the root category up to a desired depth. 
\citet{AspertEtAl:19} introduced a graph database structure ---in such management systems traversing and performing the breadth-first search is very efficient--- and provide the database for the English Wikipedia with monthly updates. 
Differently to us, these utilities, as CorpusPedia, also expect the user to input the depth up to which define the traversal for a root category.

In an approach completely unrelated to graphs, \citet{Plamada:12,Plamada:13} proposed a model for retrieving Wikipedia articles associated to a domain based on a typical search engine. Their final purpose was to retrieve parallel sentences for domain-specific statistical machine translation.
The authors processed the collection of Wikipedia articles as follows. Given two 
Wikipedia editions in language $L$ and $L'$, they
\Ni identify the subset of articles in language $L$ for which a corresponding 
article exists in language $L'$ (i.e.\ an inter-language link connects them);
\Nii index the resulting documents.
In order to retrieve the relevant articles, the index is queried with 100 
in-domain keywords ---the most frequent ones in an external in-domain corpus.
In this case, the information about the Wikipedia structure is 
not used at all and the selection of in-domain articles fully depends on their
contents. Due to the completely different nature of this system with respect to our approach, 
we use it throughout our work for comparison purposes. 
\citet{Plamada:12}  also showed the difficulties of using Wikipedia categories for 
the extraction of articles in the \emph{Alpine} domain.
In their experiments, they found that some articles within the main namespace lack a category 
tag and that the categories assigned to the same article in different languages 
do not overlap.

Full projects have been also devoted to the topic.
The ACCURAT project\footnote{\url{http://www.accurat-project.eu/}} released a toolkit for multi-level alignment and information extraction from comparable corpora. The toolkit~\cite{Pinnisetal:12} operates on different levels: \Ni alignment of comparable documents, \Nii extraction of parallel sentences, \Niii extraction of terminologies, and \Niv extraction of named entities. The toolkit can be applied on the Wikipedia to extract a general domain comparable corpus, and it retrieves the documents by analysing comparable segments in the candidates. 
A series of similarity metrics is applied to determine the level of comparability of a pair of two documents.
The approach and aim of the tool is completely different to ours. In their 
case, the main purpose is the comparability of corpora. Our focus is the domain; 
the comparability is a direct consequence given that \Ni at corpus level, if 
the languages cover the same domain, the corpora are comparable and \Nii at 
document level, comparability can be established using the inter-language 
links\footnote{Not all the articles that are comparable are linked among 
languages but a large percentage is.}.

Linguatools\footnote{\url{http://linguatools.org/}} released three Wikipedia-derived
corpora in 23 different languages. A monolingual corpus 
with more than 5 billion tokens; a comparable corpus with more than 41 million 
bilingually-aligned Wikipedia articles for 253 language pairs; and two parallel 
corpora, one with bilingual titles, extended with 
redirects and textlinks with almost 500\,M parallel segments, and the other 
one with 7\,k sentence pairs extracted from bilingual English--German quotations. 
Unfortunately, neither the tool nor the methodology for the
extraction are available in this case. Still, similarly to the corpus that can 
be obtained with the ACCURAT toolkit,
these comparable corpora do not belong to 
a specific domain but to the whole Wikipedia.

As said in the introduction, we release the \wt toolkit and 7,430 in-domain collections. \wt further allows for extracting the intersection and union of collections in multiple languages at the same time and  for the extraction of multilingual (in-domain) titles in several languages. 
Parallel titles can also be obtained with a tool%
\footnote{\url{https://github.com/clab/wikipedia-parallel-titles}}
from LTI/CMU, but we extend this functionality to go beyond only two languages.

\section{Models for Domain-based Article Selection}
\label{sec:models}

We approach the automatic extraction of domain-specific comparable corpora 
using two alternative approaches. 
Both approaches are language independent ---as far as the 
tools to perform standard preprocessing are at hand---,
and can be applied to any domain without \emph{a priori} information.

\subsection{Graph-based Model}	
\label{sub:graph-selection}

In this approach we take advantage of the user-generated categories associated to most Wikipedia articles.
As aforementioned, even if these categories are imperfect, they offer important hints on the 
domain an article belongs to.
Ideally, the categories and sub-categories should compose a category tree, and one could traverse the tree to extract the related categories hanging from a specific domain (root category).%
\footnote{Exploring Wikipedia's category graph is possible:
\url{http://en.wikipedia.org/w/index.php?title=Special:CategoryTree}.}
Nevertheless, the categories in the Wikipedia compose a densely-connected graph $G$ and the traversal is not trivial. 

Figure~\ref{fig:pyr-graph} is an example of the intrinsic difficulties inherent to WCG topology (although this particular example comes from the Wikipedia in English, similar phenomena can be observed in other editions). 
Firstly, the paths from different unrelated categories \textit{Space} and \textit{Language}, converge in common nodes early in the graph: in category \textit{Geometric measurement} at depth 2 and 7 respectively.
As a result, not only \textit{Geometric measurement} would be considered as a sub-category of both \textit{Space} and \textit{Language}, but all its descendants.
Notice also that the topic of the root category gets diluted as we go deeper into the graph and it can change to another topic. The 6th level departing from \textit{Language} in this path already talks about physics.
Secondly, $G$ contains cycles, as observed in the sequence \textit{Space} $\rightarrow$ \textit{Geometry} $\rightarrow$ \textit{Geometric measurement} $\rightarrow$ \textit{Dimension} $\rightarrow$ \textit{Space}. The exploration of the graph is therefore non-trivial.

\tikzstyle{block} = [rectangle, draw=white, fill=white, node distance=2.7cm, text width=5em, text badly centered, rounded corners, minimum height=2.4em, inner sep=0pt]
\tikzstyle{line} = [draw,->,-latex,black,line width=1pt]
\begin{figure}[t]
    \centering
    \begin{tikzpicture}[auto,draw=black,line width=1pt,scale=0.7,transform shape,font=\sffamily]

	\node [block, draw=orangeplot] (root2) {Language};
	\node [block, right of=root2] (cat11) {Philosophy of language};
	\node [block, below of=cat11, node distance=1.5cm] (cat12) {Theories of Language};
	\node [block, below of=cat12, node distance=1.5cm] (cat13) {Structuralism};
	\node [block, below of=cat13, node distance=1.5cm] (cat14) {Difference};
	\node [block, left of=cat14] (cat15) {Quantity};
	\node [block, left of=cat13] (cat16) {Physical quatities};
	\node [block, left of=cat16, draw=greenplot, text width=6.5em] (cat17) {Geometric measurement};
	\path [line] (root2) -- (cat11);
	\path [line] (cat11) -- (cat12);
	\path [line] (cat12) -- (cat13);
	\path [line] (cat13) -- (cat14);
	\path [line] (cat14) -- (cat15);
	\path [line] (cat15) -- (cat16);
	\path [line] (cat16) -- (cat17);
 	\node [block, left of=root2] (tmp) {};
 	\node [block, left of=tmp, draw=orangeplot] (root) {Space};
 	\node [block, below of=tmp, node distance=1.5cm] (cat1) {Geometry};
  	\node [block, left of=cat17] (cat2) {Dimension};
 	\path [line] (root) -- (cat1);
 	\path [line] (cat1) -- (cat17);
 	\path [line] (cat17) -- (cat2);
 	\path [line] (cat2) -- (root);
	
  \end{tikzpicture}
    \caption{Slice of the English WCG as in May 2020 departing from the categories \textit{Space} and \textit{Language}. Both graphs meet on the \textit{Geometric measurement} category at depth 2 and 7 respectively. Notice also the cycle around \textit{Space}.
    }
 \label{fig:pyr-graph}
\end{figure}
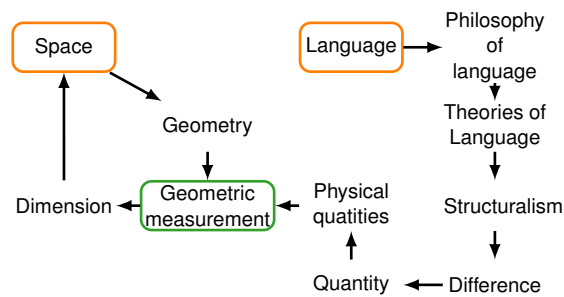

\begin{figure}
 \centering
 \includegraphics[width=\columnwidth]{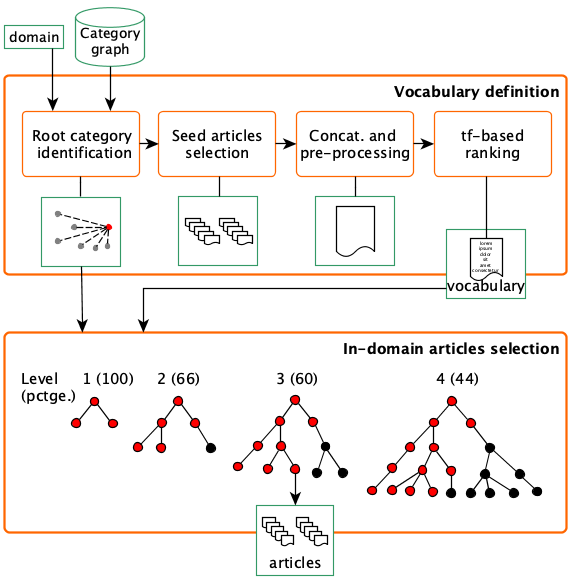}
 \caption{The two modules of the graph-based in-domain  article selection 
pipeline: vocabulary definition and articles selection. Orange rounded blocks 
represent processes. Green rectangles represent outcomes; \textit{pctge.} 
refers to the percentage of positive categories at a given tree level.
}
\label{fig:graph-pipeline}
\end{figure}

The previous example evinces that one cannot consider Wikipedia's category pseudo-tree from a root category to its leaves to define a domain. Therefore, we designed a strategy to walk through the category 
graph departing from a user-defined root category up to the level that most likely represents 
an entire knowledge domain. We tailor the Wikipedia to fit our purpose; that is, to 
build a well-formed tree representing a domain.
Figure~\ref{fig:graph-pipeline} shows the two modules of our graph-based model, which we describe below.
The input consists of the domain of interest and the pre-existing full category graph.

\paragraph{Module 1: Vocabulary Definition.} 
The objective of this module
is building the characteristic domain vocabulary $V$. It consists of four sub-modules. 
\textbf{Root category identification.} We select the root category $c_r$ that better matches the 
desired domain (e.g., \textit{Sport}). 
Our vocabulary definition and the graph exploration process 
departs from such selected node. 
\textbf{Seed articles selection.} Next, we identify every article which belongs to category $c_r$. 
The resulting set of articles is the seed for the in-domain vocabulary generation. 
If the resulting number of seed articles is small ($<10$ in our experiments) 
we include those articles associated to the children categories as well.
\textbf{Concatenation and pre-processing.}
The resulting set of articles are concatenated into one single document and 
we apply the following pre-processing operations: tokenisation, stopword 
removal, numbers, diacritics and punctuation marks removal, and stemming~\cite{Porter:80}. In order to reduce noise further, we discard tokens shorter 
than four characters (we threshold at three for Arabic as most roots in this 
language are triliteral~\cite[p.~4]{Darwish:14}). 
\textbf{Ranking.} We compute term frequency and rank the terms accordingly. 
The output of this step consists of the top-$n$ \textit{tf}-ranked terms.

\paragraph{Module 2: Graph-based Article Selection.} 
The second module explores the category graph to find 
those categories which are likely to belong to the desired domain and 
extracts the associated articles. 
The input for this step is the root category $c_r$, 
which represents the ceiling of the domain to be retrieved (e.g., 
\textit{Sport}), and the produced characteristic vocabulary $V$. We perform a 
breadth-first search departing from node $c_r$. Different criteria can be considered to stop the search in order to avoid 
exploring practically the entire graph. Our stopping criterion is inspired by 
the classification tree-breadth first search model by~\citet{Cui:08}. The 
objective is scoring the explored categories in order to assess their likelihood 
of actually belonging to the desired domain.
Our strategy assumes that a category belongs to the domain only if its 
title contains at least one of the words in the previously defined vocabulary. 
Nevertheless, many categories exist that may not include any of the words in the 
vocabulary. 
A na\"ive but efficient solution is to consider subsets of categories according 
to their depth with respect to the root, and include or exclude the full 
subset (level). 
Therefore, we traverse $G$ and score each tree level by measuring the 
percentage of its categories that are associated to the domain by means of 
containing at least one term of the vocabulary in the title. The process stops 
when less than $k\%$ of the categories are related to the vocabulary.
In the example represented in Figure~\ref{fig:graph-pipeline}, both categories 
in the first level fulfill the constraints. Two out of three do in the second level and 
three out of five do in the third one. In the fourth level only four out of nine 
categories include a characteristic term in their titles. Assuming a threshold 
of $50\%$, that level in the tree is discarded and all the articles associated 
to the categories of the tree, up to the third level, compose the output of this process.

This article selection method has two free parameters: the size of the vocabulary 
and the percentage of articles with an in-domain term in the title that we 
require to include a level in the extraction.
Section~\ref{sec:analysis} describes the characteristics of the extractions according to these parameters.

\subsection{IR-based Model}	
\label{sub:ir-selection}

For comparison with the graph-based model, we include one based on standard 
IR techniques. A model for retrieving Wikipedia articles associated to a domain based on a typical 
search engine was proposed in~\citet{Plamada:13} (see Section~\ref{sec:related}). 
Here, we implement a similar method that consists of three steps, as depicted in Figure~\ref{fig:ir-pipeline}:

\begin{figure}
 \centering
 \includegraphics[width=\columnwidth, height=0.99\columnwidth]{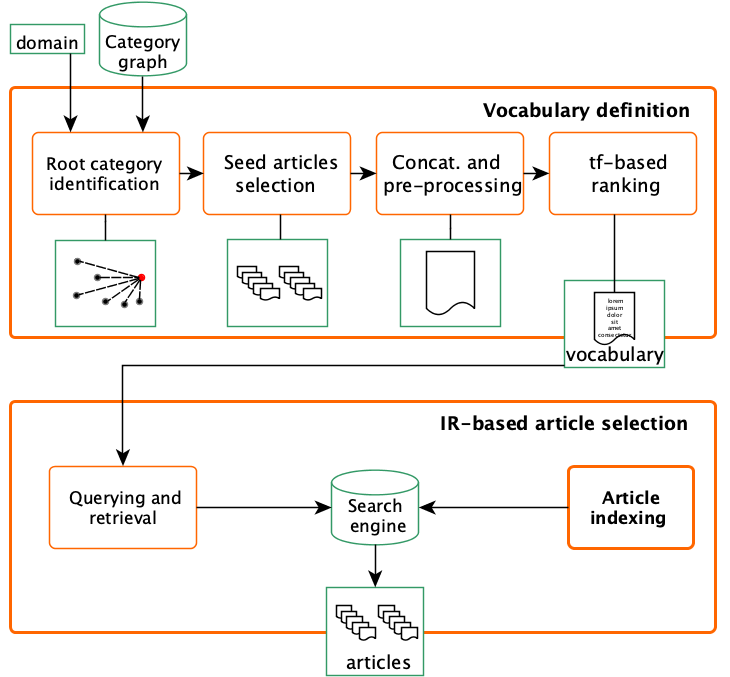}
 \caption{The IR-based in-domain  article selection pipeline. Notice that vocabulary 
definition is identical to the one in the graph-based approach (cf. 
Figure~\ref{fig:graph-pipeline}). Orange rounded blocks 
represent processes. Green rectangles represent outcomes.
}
\label{fig:ir-pipeline}
\end{figure}

\paragraph{Module 0: Article Indexing.}
As an offline preliminary process, we index every Wikipedia edition 
 and set up a search engine (right-hand side of the bottom block in Figure~\ref{fig:ir-pipeline}). 
For this, we use the Apache Lucene open-source search engine%
\footnote{\url{https://lucene.apache.org}}
and perform a pre-processing pipeline identical to the one in the graph-based model.

\paragraph{Module 1: Vocabulary Definition.}
Again, we perform the same pre-processing and ranking 
strategy to define the necessary domain vocabulary.

\paragraph{Module 2: IR-based Article Selection.}
Finally, we query the search engine with the vocabulary
and retrieve the set of articles that presumably belong to the domain 
of interest. The quality of the vocabulary is even more relevant in this case 
and a loose list could involve retrieving almost the full document collection. 

The IR-based article selection method has two free parameters, the size of the 
vocabulary and the threshold for the relevance of the articles. 
Section~\ref{sec:analysis} describes the characteristics of the extractions according to these parameters.

\section{In-Domain Collection Extraction}
\label{sec:analysis}

We explore in this section the collections obtained when applying the two described models but, before, we describe the experimental framework where they are going to be evaluated.

\subsection{Framework and Domains Definition}
\label{sec:settings}

\begin{table*}
\caption{
Statistics of the ten Wikipedia editions considered in this work in terms of number of articles and categories. 
Editions ranked according to their number of categories. The cumulative 
intersection is measured with respect to all the languages below a given row.
}
\label{tab:WPeditions}
\centering
\small
\medskip
  \begin{tabularx}{0.67\textwidth}{l@{\hspace{2em}}r@{\hspace{1.5em}}r@{\hspace{1.5em}}cr@{\hspace{2em}}rr}
   \toprule
                                &   &  & Ratio   & Intersect.  & Intersect.   \\
    Edition  & Articles   & Categories &arts/cats& categories  & articles \\
   \midrule     
    English  &   4,514,317 & 1,206,065 &   3.7 & --      &  -- \\ 
    French   &   1,487,637 &   303,156 &   4.9 & 141,994 & 933,082\\ 
    Spanish  &   1,070,407 &   261,681 &   4.1 &  72,263 & 421,008\\ 
    German   &   1,563,831 &   224,826 &   7.0 &  38,038 & 285,475\\ 
    Arabic   &     331,187 &   122,195 &   2.7 &  19,115 &  87,571\\ 
    Romanian &     255,667 &    95,657 &   2.7 &  12,776 &  39,182\\ 
    Catalan  &     435,817 &    55,099 &   7.9 &   5,467 &  31,666\\ 
    Basque   &     249,400 &    44,879 &   5.6 &   4,409 &  19,797\\ 
    Greek    &     100,703 &    30,655 &   3.3 &   3,336 &  12,539\\ 
    Occitan  &      90,270 &    15,518 &   5.8 &   2,081 &   6,811\\ 
   \bottomrule
 \end{tabularx}
\end{table*}


We select ten Wikipedia editions that serve as archetypes for different 
development levels, both in terms of amount of articles and richness of 
contents:
English, French, Spanish, German, Arabic, Romanian,  
Catalan, Basque, Greek, and Occitan. 
The set also covers different language families, including Germanic, 
Romance, and Semitic.
We use  dumps%
\footnote{\url{https://dumps.wikimedia.org} 
of the ten language editions from January and February 2015
} and preprocess them with 
JWPL~\cite{Zesch:08}\footnote{\url{https://dkpro.github.io/dkpro-jwpl/}}.

We use only the subset of content articles in the dumps ---those that belong to 
the main namespace---, and discard redirection and disambiguation pages.%
\footnote{Most of such articles are labelled as such in the dumps, but some instances 
lack any labelling. We apply some heuristics with the aim of discarding 
such unlabelled, still undesired, instances. That includes the search of 
patterns such as \texttt{\{\{numberdis\}\}} in the title or 
\texttt{\{\{disambig\}\}} in the article body.}
Table~\ref{tab:WPeditions} summarises the main figures of the resulting collections.

We define a set of root categories in order to choose the domains in our study. 
The root categories should mimic the choice of topics or domains that a user would
be interested in. Following~\citet{HechtDarreb:10}, we look for the \emph{globally relevant concepts} and assume
that a category represents a general domain if it appears in all ten languages  ---even if those ten languages do not cover all the majority cultures in the world. 
Applying this constraint results in a 
pool of $2,081$ categories (cf. Table~\ref{tab:WPeditions}).
We further eliminate categories starting with the same word, keeping 
only one of the family in any of the languages. The purpose is gathering a more 
heterogeneous and general set.%
\footnote{Given that the categories that begin with the 
same word are usually specifications of a more general category (e.g., 
{\textit{Sport}, \textit{Sport in Denmark}, \textit{Sport in Lithuania}, 
\textit{Sport in Moldova}, \textit{Sport in New Zealand}}).}
We eliminate categories that begin with a digit as well for similar reasons. 
This cleanup results in a final collection of $741$ categories.
Categories used in previous research are included ---if not 
already present--- for comparison purposes: \textit{Archaeology, Linguistics, Physics, 
Biology}, and \textit{Sport}~\cite{GamalloGonzalez:11};
\textit{Mountaineering}~\cite{Plamada:13} and 
\textit{Computer Science}~\cite{Barronetal:15}. Observe that \textit{Computer 
Science} does not exist in the Greek edition  nor \textit{Mountaineering} in the Occitan one.
With these additions, we finally consider 743 core domains.

\begin{table*}[t]
\caption{Number of articles per category used to build the domain vocabularies 
(mean $x$, standard deviation $\sigma_x$ and mode $m$) for the ten 
Wikipedia editions used and the 743 domains. Only for those 
categories with less than 10 articles in the root, the first
children are also considered. The last two columns show the number of elements of 
the vocabulary when the top 10\% of the terms are considered.}
\label{tab:rootArticles}
\centering
\small
\medskip
\begin{tabularx}{0.68\linewidth}{l rrr c  rrr c rr}
   \toprule
           & \mc{3}{c}{\# root articles} & & \mc{3}{c}{\# (root 
articles} & &\mc{2}{c}{Max. Vocabulary}\\
           &  \mc{3}{c}{  }              & & \mc{3}{c}{+children)}        & &   
 
 \mc{2}{c}{(top 10\%)}\\
           & \mc{1}{c}{$x$} & \mc{1}{c}{$\sigma_x$} & \mc{1}{c}{$m$} & & 
\mc{1}{c}{$x$} & \mc{1}{c}{$\sigma_x$} & \mc{1}{c}{$m$} & & \mc{1}{c}{$x$} & 
\mc{1}{c}{$\sigma_x$} \\
   \midrule
    English   &  99 &1,332 &  2  &  & 533 & 3,710 & 10 &  &1,154 & 2,030 \\
    French    &  45 &   75 &  7  &  & 304 & 2,710 & 13 &  &  755 & 1,336 \\ 
    Spanish   &  39 &  145 &  2  &  & 141 &   750 & 14 &  &  561 &   720 \\ 
    German    & 193 &2,104 &  2  &  & 405 & 2,502 & 10 &  &1,641 & 3,417 \\ 
    Arabic    &  46 &   76 & 10  &  &  81 &   239 & 10 &  &  461 &   488 \\ 
    Romanian  &  20 &   39 &  6  &  &  56 &   177 & 12 &  &  301 &   409 \\ 
    Catalan   &  28 &   36 & 18  &  &  87 &   527 & 12 &  &  294 &   266 \\ 
    Basque    &  17 &   90 &  2  &  &  47 &   135 & 12 &  &  187 &   225 \\ 
    Greek     &  15 &   24 &  8  &  &  42 &   182 & 10 &  &  299 &   372 \\
    Occitan   &   8 &   27 &  1  &  &  22 &    80 &  1 &  &  102 &   185 \\ 
  \bottomrule
\end{tabularx}
\end{table*}

\subsection{Nomenclature and Systems Definition}
\label{ss:naming}
From now on, WikiTailor (WT)  refers to the graph-based selection method and IR to the IR-based one. 
For WT, we analyse the collections gathered including levels with more than $50\%$ or $60\%$ of their titles within the domain (percentage of \textit{positives}), and using vocabulary sizes of the top $10\%$ of terms (WTall), the top 100 within the $10\%$ (WT100) or the top 500 also within the $10\%$ (WT500). 
These models are named individually \wtV{100}, \wtV{500}, \wtV{all}, \wtVI{100}, \wtVI{all} or, by groups using a wildcard, \wtVall, \wtVIall, \wtall{100}, \wtall{all}.

For IR, we query the engine with the top 100 or 50 terms. 
The first threshold allows for a direct comparison with~\citet{Plamada:13}. 
In their case, the characteristic vocabulary is defined as the 100 most 
frequent words (not terms) in an external corpus. Our IR model is clearly inspired 
by theirs, but we try to keep all the requirements fulfilled inside 
the Wikipedia itself, hence we avoid using external corpora.
In the experiments, we build the collection with all the retrieved articles (IRall), those with a relevance score higher than a hundredth of the maximum (IR100) or those with a relevance higher than a tenth of the maximum (IR10). The combined nomenclature is equivalent to WT models: \irC{10}, \irC{100}, \irC{all}, \irV{10}, \irV{100} for individual models and a wildcard indicates groups.

\begin{figure}
\centering
 \includegraphics[width=1.03\columnwidth,angle=0]{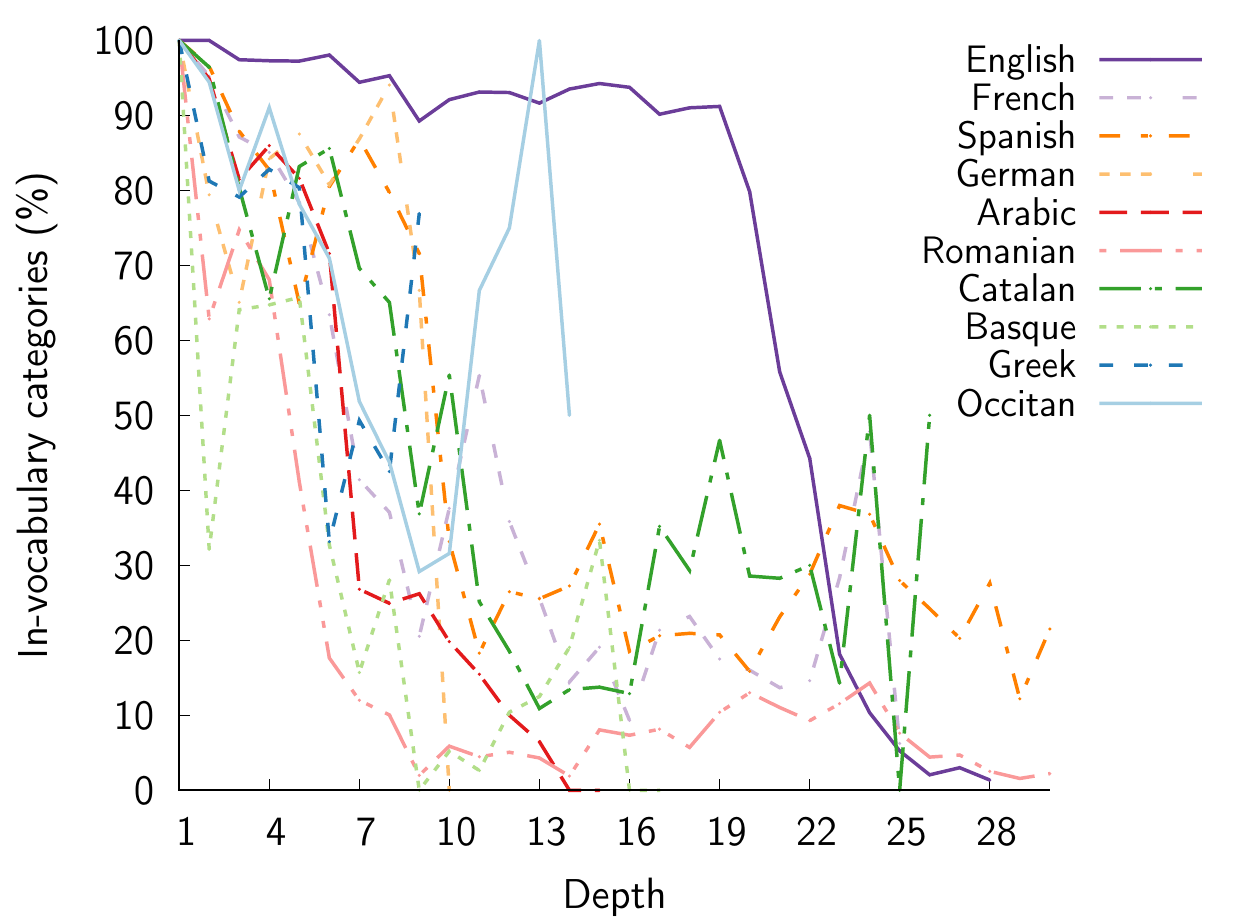}
 \includegraphics[width=1.03\columnwidth,angle=0]{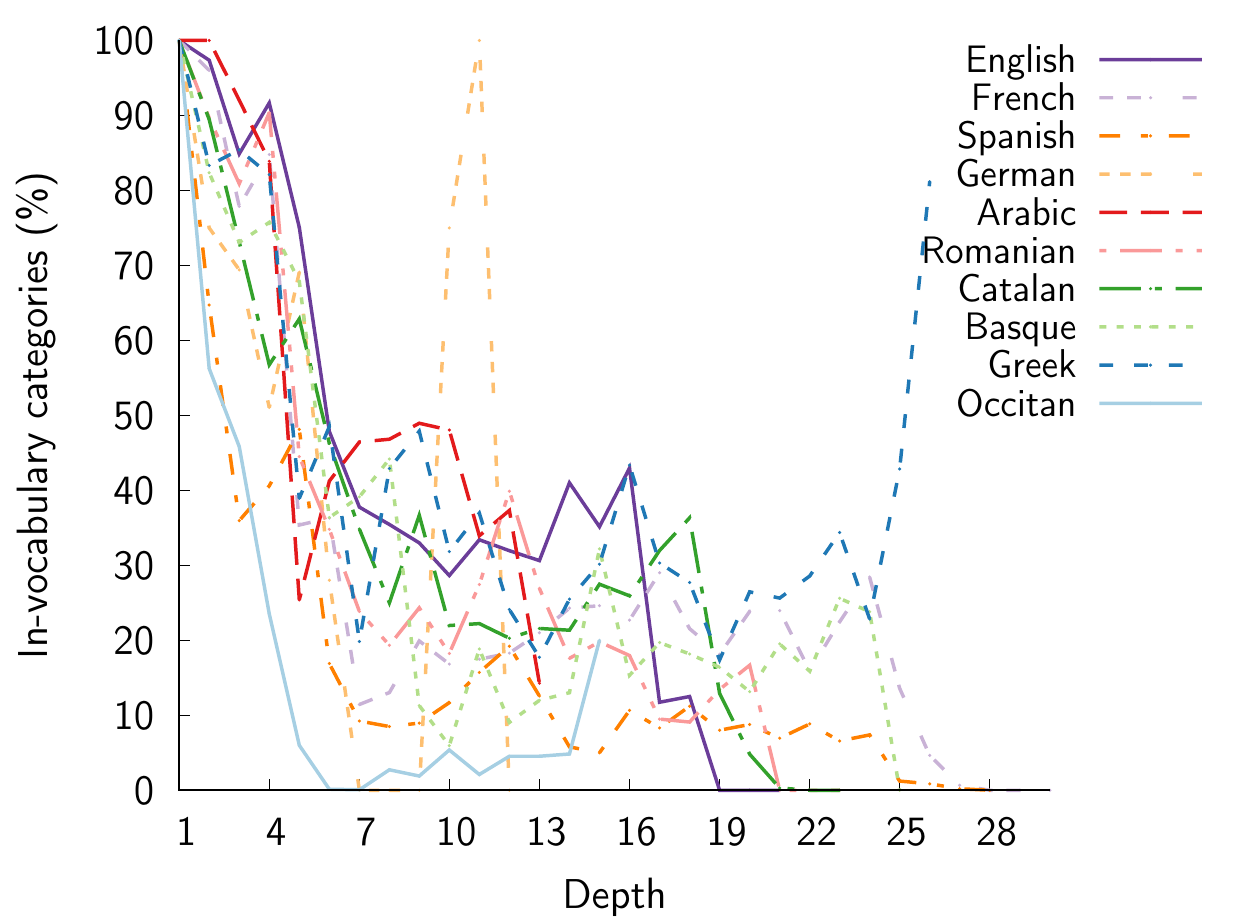}
\caption{Percentage of categories associated to the domain \textit{Sport} (top) and \textit{Astronomy} (bottom) according to the criterion described in Section~\ref{sub:graph-selection} as a function of the distance to the root category.
}
\label{fig:cat_score}
\end{figure}

\subsection{Characteristic Vocabulary}
\label{ss:vocab}

The first step in both architectures involves the extraction of the characteristic vocabulary of the domain.
Following the pipeline described in Section~\ref{sub:graph-selection}, we extract the vocabularies for different language editions in the 743 categories (domains).
Table~\ref{tab:rootArticles} shows statistics on the number of articles and size of the vocabularies. As a general trend,
the number of root articles diminishes with the size of the Wikipedia edition. 
This is true for all the languages but German and Arabic. Notice that even for 
English ---the largest edition---, the mean of root articles is 99, but the mode 
is as low as 2. Therefore, in many domains the root articles are not enough to obtain a 
large enough vocabulary. This is somehow solved by including also the articles 
in the subcategories when there are less than 10 articles in the root. 
In general, there is a chain relation: the larger the edition, the larger the
amount of articles in the root category. This results in more terms and larger 
vocabularies, potentially inducing to vocabularies with a lot of noise for 
large editions or for editions such as the German one, with lots of root 
articles. Since the quality of this vocabulary is a core factor in our methods, 
we explore several alternatives in our experiments.
Taking only the top-$10\%$ of the terms, the size of the vocabulary is completely
language-dependent. A similar thing happens with 500 elements, since the cut only 
affects major languages. For the last configuration with a maximum of 100 elements, 
the size of the vocabulary is the same, at least on average, for all the languages.

We can now study the distribution of this vocabulary along the graph.
We consider that a category belongs to the desired domain if it has an in-vocabulary term in its title.
Figure~\ref{fig:cat_score} depicts the evolution in the percentage of 
categories supposedly associated to the \textit{Sport} domain (top plot) and to
\textit{Astronomy} (bottom plot) in the ten Wikipedia editions under study.
As expected, the farther the level from the root, the lower the extent of 
associated categories (but also the larger the amount of elements). Peaks
at deeper levels can appear due to the noisy category structure of the Wikipedia, 
that 
makes that after departing from the original domain, the path returns to it
(e.g., peak at level 13th for \textit{Sport} in Occitan or at level 12th
for \textit{Astronomy} in German).
Nevertheless, the distribution is rough and, at the lowest levels, the small 
number of articles can lead to artificial canyons in the curves 
(e.g., canyon at the 2nd level for \textit{Sport} in Basque). This effect 
depends on the domain and the language. In this work, we deal with more than 
7,000 domains (743 domains times 10 languages), so, on average, the effect is 
not important and all the 
process is done fully automatically. However, in order to obtain a corpus in a 
concrete language and domain, a visual inspection of the shape of this curve 
helps to determine the stopping point of the method.

\begin{table*}[ht!]
\caption{Selected depth threshold per category (mean $x$, standard deviation $\sigma_x$ and mode $m$) for the ten Wikipedia editions used and the 743 domains.
}
\label{tab:depthCats}
\small
\centering
\medskip
  \begin{tabularx}{0.90\textwidth}{l @{\hspace{1em}} rrr rrr rrr  rrr rrr} 
   \toprule
             & \mc{3}{c}{50-WT100} & \mc{3}{c}{50-WT500} & \mc{3}{c}{50-WTall} & \mc{3}{c}{60-WT100} & \mc{3}{c}{60-WTall}\\
             \cmidrule(lr){2-4}  \cmidrule(lr){5-7} \cmidrule(lr){8-10}  \cmidrule(lr){11-13} \cmidrule(lr){14-16}
             & \mc{1}{c}{$x$} &  \mc{1}{c}{$\sigma_x$} & \mc{1}{c}{$m$} 
             & \mc{1}{c}{$x$} &  \mc{1}{c}{$\sigma_x$} & \mc{1}{c}{$m$} 
             & \mc{1}{c}{$x$} &  \mc{1}{c}{$\sigma_x$} & \mc{1}{c}{$m$} 
             & \mc{1}{c}{$x$} &  \mc{1}{c}{$\sigma_x$} & \mc{1}{c}{$m$} 
             & \mc{1}{c}{$x$} &  \mc{1}{c}{$\sigma_x$} & \mc{1}{c}{$m$} \\
   \midrule    
    English  &  5.9 & 2.8 & 5 & 9.6 & 7.3 & 8 &12.4 & 11.1 & 8 & 5.2 & 2.4 & 7 &10.5 & 9.6 & [7, 8]\\
    French   &  4.3 & 1.9 & 5 & 5.1 & 2.3 & 5 & 5.7 &  3.6 & 5 & 3.8 & 1.9 & 5 & 5.2 & 3.2 & 5 \\
    Spanish  &  4.4 & 2.1 & 2 & 6.0 & 3.7 & 2 & 6.9 &  5.8 & 2 & 3.8 & 1.9 & 2 & 5.8 & 4.4 & [2, 6]\\
    German   &  3.4 & 1.9 & 2 & 3.8 & 2.1 & 2 & 4.0 &  2.2 & 2 & 3.1 & 1.8 & 2 & 3.8 & 2.2 & 2 \\
    Arabic   &  3.6 & 2.3 & 1 & 4.7 & 3.7 & 1 & 6.1 &  4.6 & 5 & 2.9 & 2.0 & 1 & 5.2 & 3.4 & 5 \\
    Romanian &  3.4 & 1.8 & 2 & 3.8 & 2.1 & 2 & 3.8 &  2.1 & 2 & 3.2 & 1.6 & 2 & 3.6 & 2.0 & 2 \\
    Catalan  &  3.3 & 1.9 & 2 & 3.8 & 2.2 & 2 & 3.8 &  2.3 & 2 & 2.9 & 1.8 & 2 & 3.4 & 2.1 & 2 \\
    Basque   &  3.1 & 1.5 & 2 & 3.3 & 1.7 & 2 & 3.3 &  1.7 & 2 & 2.8 & 1.4 & 2 & 3.1 & 1.6 & 2\\
    Greek    &  3.0 & 1.6 & 2 & 3.3 & 1.8 & 2 & 3.3 &  1.9 & 2 & 2.8 & 1.5 & 2 & 3.1 & 1.8 & 2\\
    Occitan  &  2.4 & 1.3 & 2 & 2.5 & 1.4 & 2 & 2.5 &  1.4 & 2 & 2.2 & 1.2 & 2 & 2.4 & 1.3 & 2 \\
   \bottomrule  
  \end{tabularx}
\end{table*}

\begin{table*}[t]
\caption{Mean $N$ and standard deviation $\sigma_N$ of the number of articles per domain for the WikiTailor model (top) and the IR-based model (bottom). We 
show five systems with different values for the two free parameters in both cases (cf.\ Section~\ref{sec:models} for a description). Left-most numbers indicate the 
ranking of the edition in number of articles (cf. Table~\ref{tab:WPeditions}).
}
\label{tab:collections}
\centering
\small
\medskip
  \begin{tabularx}{1.0\textwidth}{lr@{\hspace{0.7em}}rr@{\hspace{0.7em}}rr@{\hspace{0.7em}}rr@{\hspace{0.7em}}rr@{\hspace{0.7em}}r} 
   \toprule
 & \mc{2}{c}{50-WT100} & \mc{2}{c}{50-WT500} & \mc{2}{c}{50-WTall} & \mc{2}{c}{60-WT100} & \mc{2}{c}{60-WTall}\\
             \cmidrule(lr){2-3}  \cmidrule(lr){4-5} \cmidrule(lr){6-7}  \cmidrule(lr){8-9} \cmidrule(lr){10-11}
 & \mc{1}{c}{$N$} & \mc{1}{c}{$\sigma_N$}  
             & \mc{1}{c}{$N$} & \mc{1}{c}{$\sigma_N$}  
             & \mc{1}{c}{$N$} & \mc{1}{c}{$\sigma_N$}   
             & \mc{1}{c}{$N$} & \mc{1}{c}{$\sigma_N$}   
             & \mc{1}{c}{$N$} & \mc{1}{c}{$\sigma_N$}  \\
   \midrule    
    1 English  & 50,514 &121,881 & 513,615 & 1,170,041 &1,008,340 & 1,780,484 & 27,903 & 59,838 &734,168 & 1,544,054  \\  
    3 French   &  8,278 & 26,483 &  18,518 &    79,134 &   40,207 &   182,940 &  5,869 & 21,342 & 29,717 & 147,592  \\ 
    4 Spanish  &  6,638 & 17,050 &  34,556 &    97,268 &   63,450 &   166,908 &  4,463 & 14,257 & 36,445 & 120,180 \\ 
    2 German   &  2,752 &  9,573 &   4,131 &    16,658 &    5,150 &    19,860 &  2,199 &  8,948 &  4,671 & 19,184  \\
    6 Arabic   &  2,999 &  9,546 &  20,441 &    58,483 &   36,969 &    87,961 &  1,541 &  6,354 & 20,456 & 60,213  \\
    7 Romanian &  1,398 &  8,683 &   2,078 &    11,875 &    2,396 &    13,649 &    796 &  3,590 &  1,766 & 11,207  \\
    5 Catalan  &  1,140 &  4,693 &   2,041 &     9,891 &    2,319 &    10,991 &    686 &  2,317 &  1,601 &  7,889  \\
    8 Basque   &    440 &  1,654 &     907 &     4,673 &    1,130 &     6,638 &    320 &  1,215 &    748 &  4,207  \\
    9 Greek    &    390 &  2,008 &     777 &     4,218 &      833 &     4,649 &    298 &  1,407 &    679 &  4,097  \\
    10 Occitan &    104 &    598 &     247 &     2,480 &      308 &     3,051 &     68 &    374 &    114 &    774  \\
   \bottomrule  

  \end{tabularx}
   \vspace*{1cm}\\
  \begin{tabularx}{1.0\textwidth}{lrrrrrrrrrr} 
   \toprule
   & \mc{2}{c}{100-IR10} & \mc{2}{c}{100-IR100} & \mc{2}{c}{100-IRall}  & \mc{2}{c}{50-IR10} & \mc{2}{c}{50-IR100}  \\
             \cmidrule(lr){2-3}  \cmidrule(lr){4-5} \cmidrule(lr){6-7}  \cmidrule(lr){8-9} \cmidrule(lr){10-11}
    & \mc{1}{c}{$N$} & \mc{1}{c}{$\sigma_N$}  
             & \mc{1}{c}{$N$} & \mc{1}{c}{$\sigma_N$}  
             & \mc{1}{c}{$N$} & \mc{1}{c}{$\sigma_N$}   
             & \mc{1}{c}{$N$} & \mc{1}{c}{$\sigma_N$}   
             & \mc{1}{c}{$N$} & \mc{1}{c}{$\sigma_N$}  \\
   \midrule    
    1 English  & 64,239 & 73,248 & 1,119,637 & 482,339 &3,947,077&221,129      & 52,030 & 69,135 & 976,547 & 488,533       \\
    3 French   & 18,158 & 17,871 &   331,936 & 145,261 &1,235,344& 88,852      & 15,258 & 18,115 & 308,093 & 152,543       \\
    4 Spanish  & 21,490 & 19,605 &   314,612 & 105,193 & 958,399 & 50,521      & 17,791 & 20,454 & 283,654 & 113,708       \\
    2 German   & 12,887 & 18,876 &   378,195 & 218,271 &1,434,164& 70,250      & 12,843 & 21,628 & 344,392 & 225,517       \\
    6 Arabic   &  4,622 &  4,082 &    61,882 &  25,026 & 274,589 & 36,760      &  4,188 &  4,020 &  58,284 &  24,957       \\
    7 Romanian &  1,750 &  1,839 &    33,018 &  15,213 & 162,608 & 32,409      &  1,567 &  1,882 &  32,199 &  16,364       \\
    5 Catalan  &  5,959 &  5,058 &    99,703 &  38,195 & 370,917 & 35,443      &  5,601 &  5,940 &  97,653 &  39,756       \\
    8 Basque   &  1,819 &  2,732 &    33,173 &  21,743 & 139,099 & 33,196      &  1,741 &  2,889 &  33,421 &  23,174       \\
    9 Greek    &  2,547 &  2,378 &    40,074 &  14,550 &  97,417 & 46,153      &  2,639 &  3,349 &  41,396 &  16,108       \\
    10 Occitan &    419 &  2,040 &     6,397 &   7,052 &  42,554 & 21,270      &    487 &  2,331 &   6,617 &   7,125       \\
   \bottomrule  
  \end{tabularx}
\end{table*}  

\subsection{Collections Characteristics}
\label{ss:analysis}
WikiTailor determines automatically the depth from the root up to which it should extract articles according to the percentage of in-vocabulary categories, and this is a crucial point for the extraction. Different percentages lead to different stopping points and consequently different collection sizes. 
Looking at the specific numbers in Table~\ref{tab:depthCats}, we see that the \textbf{threshold depth} seems to be directly proportional to the size of the characteristic vocabulary and the number of categories. In general, the more categories in a Wikipedia edition, the more levels are used to describe a root category. 
These two features are more important than the alternatives of taking levels with a $50\%$ or a $60\%$ of positives. 
For a given language, the most relevant feature is the size of the vocabulary, specially for small editions: smaller vocabularies imply smaller threshold depths.
For Romanian, Catalan, Basque, and Greek systems with $50\%$ of positives select a mean boundary depth of 3 for 
WT100 and 4 for WTall. The change is less significant for the systems with $60\%$. However, if one considers large editions, the change is striking in both 
cases. In English, systems with $50\%$ of positives select a mean threshold depth of 6 for WT100 and 12 for WTall 
(5 and 11 for the $60\%$ systems). 
So, for the editions with more articles, we 
also extract all the articles from a larger sub-tree, and that favours even more 
the extraction of huge in-domain corpora for English and more modest ones for 
the other languages. As before, Arabic and German seem to be out of place. If we 
rank the editions according to the number of categories, Arabic has a higher 
than expected mean selected depth per domain, given its position in the ranking. German has it lower.
All differences among languages are reduced for small and similar vocabularies (\wtall{100}).

The top rows of Table~\ref{tab:collections} show the \textbf{size of the collections}
extracted with the WT model. The size for every system and language is a 
direct consequence of the aforementioned. Except for Arabic and German, the 
larger the edition, the larger the extracted collection of in-domain articles, 
but for small vocabularies differences among languages are less extreme. The 
loss in number of articles in English for small vocabularies respect to \wtall{all}\ is 
remarkable (from 1\,M in \wtV{all}\ to 50\,k in \wtV{100}). This is not the case for 
German (5\,k vs. 3\,k) although its initial vocabulary for \wtV{all}\ was even larger 
than the English one.

The bottom rows of Table~\ref{tab:collections} describe the in-domain 
corpora extracted with the IR model. In general, IR retrieves larger collections 
than WT, up to the point that for queries with 100 terms and without any 
threshold for the relevance score (IRall) the extracted corpus can be almost the 
full Wikipedia.
Also notice that the number of extracted articles in the IR 
models is proportional to the size of the collection instead of to the number of 
categories, as it happens with WT and small vocabularies (numbers on the 
left of the language in the table rank the editions according to the number of 
articles, the actual order in the table is regarding the number of categories).
As expected, queries with less elements (\irVall\ vs. \irCall) retrieve smaller collections. 
Some exceptions appear for Basque and Greek. This occurs when one does 
not look at the collection with all the hits (IRall) but at those recovering a 
percentage of the maximum score. Since the maximum score changes when using 100 
query terms and 50 query terms, the same can happen for the number of elements.

Our two methods, WT and IR,  build very different corpora ---specially in \textbf{content}. WT collections, which are 
smaller, are not a subset of the IR ones.%
\footnote{Except in the cases in which IRall is 
the reference, the system that selects almost the whole Wikipedia for a 
given domain.}
For example, if one compares \wtV{100} and \irC{10},
two similar collections in terms of size, only between $20-60\%$ 
of the WT articles and a $5$--$15\%$ of the IR ones appear in 
the intersection between the corresponding extractions by both models. 
The common articles cover a larger percentage of the WT collections 
because their size is smaller. The ranges in the previous figures, describe the 
behaviour for the different languages. Large editions 
have a lower percentage of common articles (for example $23\%$ and $56\%$ for 
WT in English and Greek respectively, and $8\%$ and $4\%$ for IR in the 
same languages). Previous to the evaluation of the quality of the 
collections, a collection built from the union of the different systems seems a 
way to enlarge the amount of data; specially for small editions, where it can 
be more useful.

As a final remark, notice that these results correspond to the monolingual scenario. 
A multilingual comparable corpus is just the set of collections of the same domain for each language.
We can increase the degree of comparability \cite{GamalloGonzalez:11,SuBabych:12}
by selecting a subset of equivalent articles 
in a straightforward way thanks to Wikipedia's inter-language links. 
Once the monolingual corpora have been retrieved, the union or intersection of their 
linked articles builds the final comparable corpus for the desired domain and 
languages.

\subsection{Comparison to Similar Systems}
\label{ss:comparison}

\citet{GamalloGonzalez:10,GamalloGonzalez:11} obtained comparable corpora in 
Spanish, English and Portuguese in the domains of \textit{Archaeology, 
Linguistics, Physics, Biology}, and \textit{Sport} based also on Wikipedia's 
categorisation. The comparison with our model is difficult because the Wikipedia 
edition used differs in six years 
and the editions have doubled its size during this period. Besides, they report the size of their comparable corpora in MB and not in number of articles. The single comparison we can do is that for the comparable corpus obtained for \textit{Archaeology} in English and Spanish. Their most flexible (tight) method was able to retrieve 1120 (34) articles in English and 462 (34) in Spanish. 
In our case with a different dump, the most flexible method (WPall with $50\%$ of positives) goes up to depth 16 and retrieves almost the full Wikipedia for English (4,442,585 articles) and reaches depth 11 in Spanish with a total of 636,850 articles. The most restrictive method (WP100 with $60\%$ of positives) goes up to depth 5 and retrieves 65,343 articles for English and gets depth 2 with 553 articles for Spanish. A conservative model (WP100 with $50\%$ of positives) retrieves 236,951 articles in English (depth 6) and 17,335 in Spanish (depth 5). Of course, the 
accuracy of CorpusPedia will be much higher, but for some tasks the size of the 
corpus would not be enough.
Notice that at this point, we are talking about the size of the collections and not about their quality.

\citet{Plamada:13} used a very similar method to IR to extract parallel articles in the \emph{Alpine} domain for German and French. We can compare their results with the ones we have for \textit{Mountaineering} with our IR model but, again, the Wikipedia editions differ. 
They index only aligned documents according to the inter-language links, since their main purpose is to extract parallel sentences and they assume they are mostly found in aligned (parallel) documents.
Their methodology retrieves 40,000 parallel articles while our most flexible version with the same number of terms for the query (IRall with 100 term queries) retrieves almost the full Wikipedia (1,182,465 French articles 1,460,036 German articles). The conservative version (IR100 with 100 term queries) retrieves 225,422 in-domain French articles and 305,200 German ones. We can extract the subset of parallel articles from this comparable corpus via the intersection or the union of the articles. For the intersection, we use the articles that have been identified as in-domain simultaneously in German and French. For the union, we expand the  set of articles to include all the articles that have been identified as in-domain articles in one of the languages with the equivalent article in the other one in case it exists. Using the intersection, we obtain a high precision/low recall parallel set with 55,551 articles and with the union we gather a low precision/high recall corpus with 205,913 articles.

\subsection{Manual Evaluation}
\label{sub:manual}

We have generated several in-domain document collections, but we have not determined how well these documents represent the domain. In this section, we are interested in determining whether the documents in a corpus belong to a particular domain or not.
For this manual study, we select two representative systems: \wtV{100}\ and \irC{10}
and manually judge their articles in three domains in all ten languages: \emph{Astronomy}, \emph{Software}, and \emph{Sport}. 
The evaluation set 
for each language, domain and system consists of 200 articles: 100 articles exclusive to each system and 100 
articles in common to both. The articles are extracted evenly in its subset. In three cases, the number of articles in the collection is smaller than 200 and so is the evaluation set (see Table~\ref{tab:cf-precision}).

We manually annotate the 8,600 articles with three assessments each. We use the Figure Eight%
\footnote{\url{https://www.figure-eight.com/}}
platform to crowdsource this task. All the details on setting up the experiment and instructing the Turkers are in Appendix~\ref{app:crowdflower}.

\begin{table*}[t]
\caption{Results of the manual evaluation. The number of articles selected for the manual assessments are shown in Set$_{\rm WT}$ and Set$_{\rm IR}$. ``Complete set'' shows the precision obtained under the hard and soft criteria for the 50-WT100 (WT) and 100-IR10 (IR) systems. ``100-element subset'' analyses the distribution of the sets (see text). The last column shows the inter-annotator agreement measured by the Fleiss' kappa.}
\label{tab:cf-precision}
\centering
\medskip
\small
\begin{tabularx}{0.97\textwidth}{l rr rrrr rr rr rr r}
\toprule
          &  &  & \multicolumn{4}{c}{Complete set} & \multicolumn{6}{c}{100-element subset}& \\
           \cmidrule(lr){4-7} \cmidrule(lr){8-13}
          & Set$_{\rm WT}$ & Set$_{\rm IR}$ & 
          \multicolumn{2}{c}{WT} & \multicolumn{2}{c}{IR}& \multicolumn{2}{c}{WT$_{\rm only}$} & \multicolumn{2}{c}{IR$_{\rm only}$} & \multicolumn{2}{c}{$\cap_{\rm only}$}   & $\kappa_{\rm Fleiss}$ \\
          &  &  & hard  & soft         & hard  & soft         & hard  & soft       & hard  & soft         & hard  & soft &  \\   
\midrule
\emph{ASTRONOMY}     \\
English   & 200 & 200 & 0.67 & 0.82 & 0.37 & 0.44 & 0.64 & 0.84 & 0.05 & 0.07 & 0.70 & 0.81 & 0.789 \\    
French    & 200 & 200 & 0.71 & 0.81 & 0.40 & 0.45 & 0.69 & 0.80 & 0.07 & 0.07 & 0.74 & 0.83 & 0.828 \\    
Spanish   & 200 & 200 & 0.83 & 0.94 & 0.45 & 0.55 & 0.83 & 0.87 & 0.07 & 0.10 & 0.83 & 1.00 & 0.834 \\    
German    & 200 & 200 & 0.77 & 0.92 & 0.76 & 0.87 & 0.72 & 0.88 & 0.71 & 0.79 & 0.82 & 0.96 & 0.524 \\    
Arabic    & 200 & 200 & 0.59 & 0.65 & 0.41 & 0.45 & 0.37 & 0.46 & 0.02 & 0.05 & 0.81 & 0.85 & 0.816 \\    
Romanian  & 200 & 200 & 0.76 & 0.82 & 0.41 & 0.47 & 0.71 & 0.72 & 0.01 & 0.02 & 0.81 & 0.93 & 0.883 \\    
Catalan   & 200 & 200 & 0.73 & 0.85 & 0.46 & 0.51 & 0.66 & 0.79 & 0.11 & 0.11 & 0.81 & 0.92 & 0.786 \\    
Basque    & 200 & 200 & 0.95 & 0.99 & 0.48 & 0.52 & 0.96 & 0.98 & 0.03 & 0.05 & 0.94 & 1.00 & 0.929 \\    
Greek     & 200 & 200 & 0.70 & 0.74 & 0.44 & 0.48 & 0.52 & 0.57 & 0.00 & 0.04 & 0.88 & 0.92 & 0.893 \\    
Occitan   & 132 & 139 & 0.66 & 0.71 & 0.35 & 0.40 & 0.59 & 0.65 & 0.16 & 0.21 & 0.85 & 0.90 & 0.845 \\    
\midrule                                            
\emph{SOFTWARE}     \\                              
English   & 200 & 200 & 0.73 & 0.92 & 0.42 & 0.56 & 0.69 & 0.91 & 0.07 & 0.19 & 0.78 & 0.94 & 0.657 \\    
French    & 200 & 200 & 0.77 & 0.91 & 0.42 & 0.54 & 0.82 & 0.92 & 0.12 & 0.18 & 0.72 & 0.91 & 0.735 \\    
Spanish   & 200 & 200 & 0.80 & 0.93 & 0.48 & 0.56 & 0.75 & 0.93 & 0.12 & 0.17 & 0.85 & 0.95 & 0.756 \\    
German    & 200 & 200 & 0.72 & 0.84 & 0.52 & 0.65 & 0.63 & 0.80 & 0.24 & 0.41 & 0.81 & 0.90 & 0.609 \\    
Arabic    & 200 & 200 & 0.41 & 0.61 & 0.30 & 0.47 & 0.26 & 0.42 & 0.04 & 0.15 & 0.56 & 0.79 & 0.614 \\    
Romanian  & 200 & 200 & 0.84 & 0.97 & 0.49 & 0.58 & 0.84 & 0.98 & 0.13 & 0.19 & 0.85 & 0.97 & 0.788 \\    
Catalan   & 200 & 200 & 0.80 & 0.93 & 0.50 & 0.60 & 0.77 & 0.93 & 0.18 & 0.26 & 0.83 & 0.94 & 0.724 \\    
Basque    & 200 & 200 & 0.86 & 0.96 & 0.48 & 0.53 & 0.83 & 0.95 & 0.08 & 0.09 & 0.89 & 0.98 & 0.852 \\    
Greek     & 200 & 200 & 0.87 & 0.96 & 0.51 & 0.56 & 0.86 & 0.95 & 0.13 & 0.15 & 0.89 & 0.97 & 0.928 \\    
Occitan   &  17 &  73 & 0.72 & 0.89 & 0.30 & 0.58 & 0.82 & 0.91 & 0.27 & 0.55 & 0.67 & 1.00 & 0.424 \\    
\midrule                                                                                            
\emph{SPORTS}     \\                                                                                
English   & 200 & 200 & 0.70 & 0.79 & 0.44 & 0.48 & 0.55 & 0.64 & 0.01 & 0.01 & 0.87 & 0.95 & 0.897 \\      
French    & 200 & 200 & 0.92 & 0.93 & 0.50 & 0.50 & 0.86 & 0.88 & 0.02 & 0.02 & 0.98 & 0.98 & 0.947 \\      
Spanish   & 200 & 200 & 0.86 & 0.90 & 0.49 & 0.50 & 0.77 & 0.83 & 0.03 & 0.03 & 0.95 & 0.97 & 0.948 \\      
German    & 200 & 200 & 0.55 & 0.75 & 0.56 & 0.66 & 0.66 & 0.82 & 0.55 & 0.64 & 0.47 & 0.51 & 0.710 \\      
Arabic    & 200 & 200 & 0.37 & 0.44 & 0.13 & 0.16 & 0.51 & 0.58 & 0.02 & 0.02 & 0.24 & 0.31 & 0.877 \\      
Romanian  & 200 & 200 & 0.73 & 0.78 & 0.37 & 0.41 & 0.72 & 0.74 & 0.01 & 0.01 & 0.74 & 0.82 & 0.920 \\      
Catalan   & 200 & 200 & 0.60 & 0.68 & 0.33 & 0.38 & 0.56 & 0.61 & 0.02 & 0.02 & 0.65 & 0.75 & 0.876 \\      
Basque    &  29 & 111 & 0.87 & 0.97 & 0.10 & 0.12 & 0.94 & 1.00 & 0.02 & 0.02 & 0.82 & 1.00 & 0.943 \\      
Greek     & 200 & 200 & 0.91 & 0.96 & 0.44 & 0.48 & 0.94 & 0.96 & 0.00 & 0.01 & 0.89 & 0.96 & 0.841 \\      
Occitan   & 200 & 200 & 0.74 & 0.77 & 0.56 & 0.57 & 0.50 & 0.54 & 0.14 & 0.14 & 0.99 & 1.00 & 0.946 \\      
\bottomrule
\end{tabularx}
\end{table*}
 
Table~\ref{tab:cf-precision} shows the manually-judged precision results.
We calculate the precision of the extracted collections under two circumstances: \Ni \textbf{hard precision} when there is full agreement in assigning a domain among the three annotators and \Nii \textbf{soft precision} when an article is assigned to a domain by the two out of three annotators. 
For the three domains, the quality of the WT extractions is much better than those with IR\@. Even in the hard-precision setting, the mean value is 0.74$\pm$0.14 for WT and 0.43$\pm$0.12 for IR, and values per domain are close to this value.
The average values for soft precision go up to 0.84$\pm$0.13 for WT and 0.50$\pm$0.14 for IR\@. 
Focusing in the language factor,
the IR system does specially well for German, suggesting that the quality of the extracted characteristic vocabulary is better. This is an indication that the quality of the characteristic vocabulary is less important in the WT models than in the IR ones, as WT averages among all the articles in a level before extracting the whole level. On the other hand, 
WT's weakest performance comes with Arabic,
with a mean soft precision over domains of 0.57$\pm$0.11. Arabic collections are built after considering a low depth (3.6$\pm$2.3 with a mode as low as 1; cf.\ Table~\ref{tab:depthCats}). Nevertheless, the three evaluated domains are built upon a higher depth (5 for \emph{Astronomy}, 8 for \emph{Software}, and 6 for \emph{Sport}) meaning that perhaps too many articles are extracted increasing the coverage but damaging the precision. The outcome is still better than for its IR counterpart.

The difference between the WT and IR systems becomes more evident when looking into the distribution of their resulting collections. As explained before, we have built the subsets to evaluate by assuring that half of the articles in a collection are common in both systems and the other half is exclusive to each of them. That allows us not only to save in manual assessments, but also to have a clear idea of the distribution of the articles in a collection. The third vertical block of Table~\ref{tab:cf-precision} ``100-element subset'' shows the results. As expected, the articles that are common to both systems ($\cap_{\rm only}$) are those with the highest precision (0.79$\pm$0.15 for hard precision and 0.89$\pm$0.15 for soft precision on average). The quality of the articles extracted only by the WT system (WT$_{\rm only}$) are very close in quality with an average of 0.70$\pm$0.17 for hard precision and  0.80$\pm$0.17 for soft precision. The precision values are very low for articles only retrieved by the \irC{10}\ system (mean of 0.11$\pm$0.16 for hard precision and 0.16$\pm$0.20 for soft precision). The only exception is again German, where the IR$_{\rm only}$ subcollection has a hard precision of 0.50$\pm$0.24 and a soft precision of 0.61$\pm$0.19.

The last column of Table~\ref{tab:cf-precision} shows the interannotator agreement for more than two raters, the Fleiss' kappa ($\kappa_{\rm Fleiss}$)~\cite{fleiss:1971}. 
Turkers agreed the most when discriminating between \emph{Sport} and \emph{other} domain, with an average $\kappa= 0.88\pm0.07$. 
The lowest agreements occurred in the \emph{Software} domain: $\kappa= 0.74\pm0.11$. \emph{Astronomy} lies in the middle with 0.81$\pm$0.12.
Regarding the language dimension, annotators of Basque agreed the most, with $\kappa= 0.91\pm0.05$. Instances in German were the least agreed upon, with a low $\kappa= 0.61\pm0.09$. Individually, annotators of Spanish instances \textit{Sports} vs \textit{other} obtained the highest agreement: $0.95$. The lowest agreement was obtained for \textit{Astronomy} vs \textit{other} in German: only $0.52$.

Notice that for most of the evaluations (28 out of 30) we obtain either \emph{substantial agreement} ($0.61$$<$$\kappa$$<$$0.80$) or \emph{almost perfect agreement} ($0.81$$<$$\kappa$$<$$1.00$) as defined in \citet{landisKoch:1977}, and we can conclude that system \wtV{100}\ is significantly better than \irC{10}.
However, a manual evaluation is always expensive and one would like to be able to quantify automatically how adequate is a collection with respect to the desired domain for each experiment. Next section introduces the concept of domainness and addresses the issue.

\section{\textbf{\textit{Domainness}} Characterisation}
\label{sec:evaluation}

We are still interested in determining whether the documents in a collection belong to a particular domain or not. Nevertheless, describing corpora is a difficult and subjective task and the answer should not be binary, but a continuous score, especially if it is quantified automatically. 
Here, we define domainness as the degree of cohesion and representativity of a corpus with respect to a domain:
\begin{equation}
 {\rm domainness} = {\rm representativity} + {\rm cohesion} \nonumber
\end{equation}

\begin{figure}
 \centering
 \includegraphics[width=0.3\textwidth]{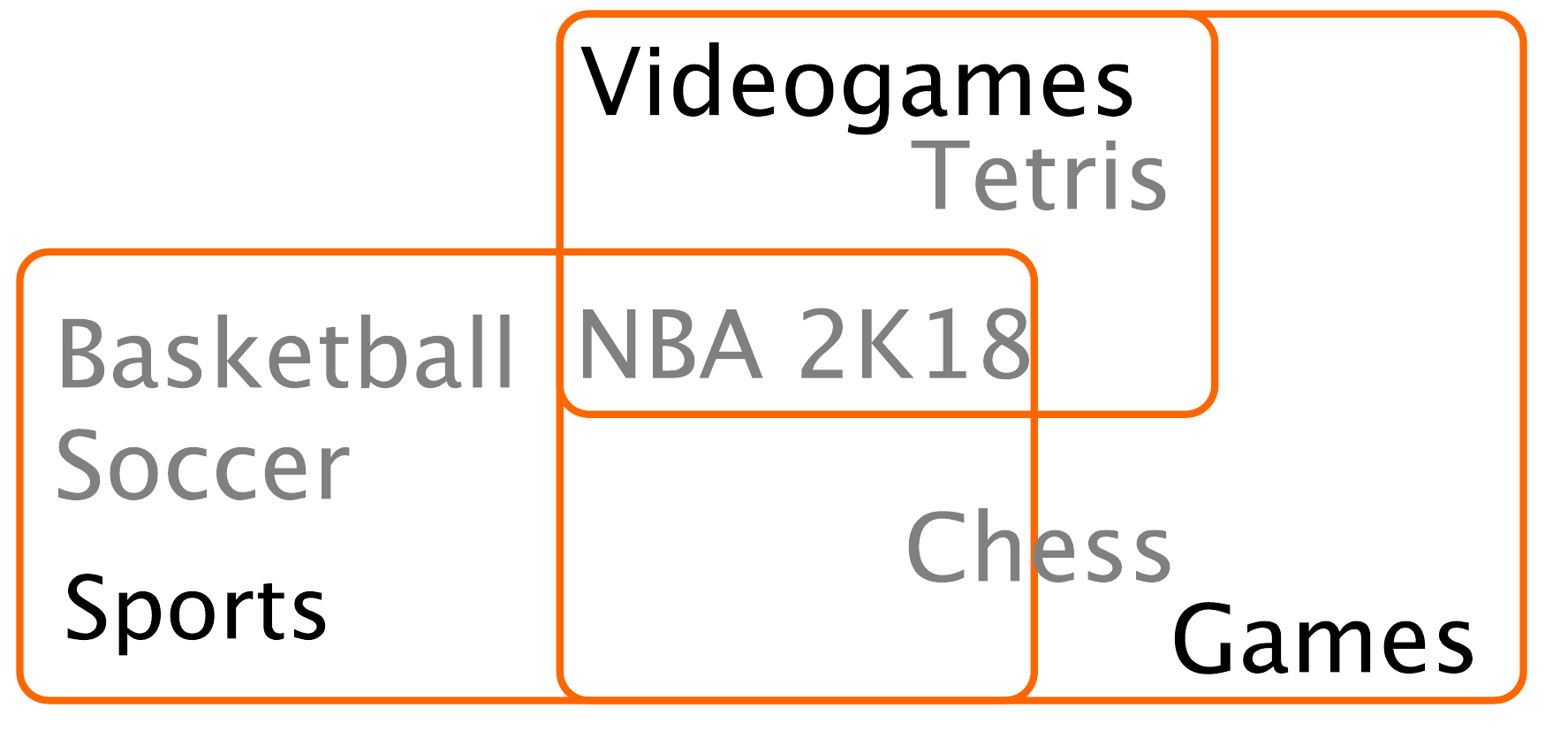}
 \caption{Example of three intersecting domains: \emph{Sport}, \emph{Games}, \emph{Videogames} (orange boxes) and articles within them (in gray).}
\label{fig:wikiexamples}
\end{figure}

\subsection{Concept Intuition}
\label{ss:intuition}

The idea behind the definition of domainness builds on the intuition that a collection should be heterogeneous but cohesive at the same time.
For illustrative purposes, Figure~\ref{fig:wikiexamples} shows three domains and five Wikipedia articles within them.
Article \textit{Basketball} clearly belongs to domain \emph{Sport}, whereas \textit{Tetris} clearly does not.
Articles such as \textit{NBA 2K18} lie within all ---\emph{Sport}, \emph{Games} and \emph{Videogames} domains--- as it represents them all.
Yet the membership of article \textit{NBA 2K18} to the \emph{Sport} domain is subjective, unless a more detailed description of the domain is given. At collection level, a collection with the previous three documents is less representative of \emph{Sport}, than a collection including articles \emph{Basketball}, \emph{Soccer} and \emph{Chess} which are more \textbf{cohesive}. Again, at what extent remains subjective ---we need a measure to quantify the difference. 

\begin{figure}
 \centering
 \includegraphics[width=0.4\textwidth]{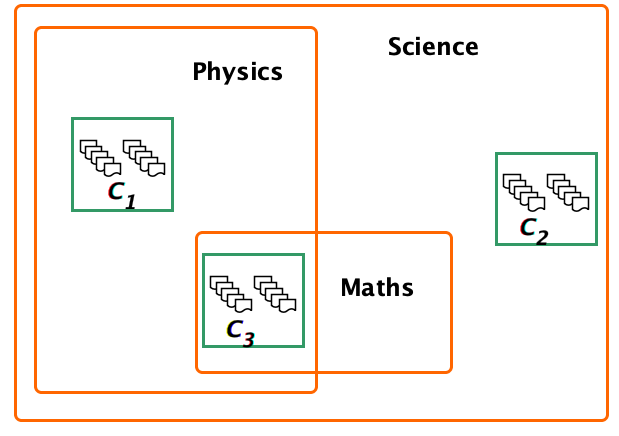}
 \medskip
 \caption{Illustrative example of \emph{representativity} and \emph{cohesion} to define the domainness of a collection of documents $C_i$. $C_1$ has the highest domainness for \emph{Physics}, whereas $C_3$ and $C_2$ have higher domainness for \emph{Science} since they have a major representativity. 
 }
\label{fig:domainness}
\end{figure}

Figure~\ref{fig:domainness} shows another example to illustrate the concept of \textbf{representativity} within a collection. Whereas collections $C_1$ and $C_3$ correspond to the \emph{Physics} domain, $C_1$ should receive a higher domainness score because articles seem to be purely about physics ($C_3$ contains articles in the intersection of physics and math). However, when measuring the domainness of the collections with respect to the \emph{Science} domain, $C_3$ should have a higher value because it has more diversity, 
i.e.\ it holds a higher representativity of the domain. From this configuration, one cannot say which of $C_2$ or $C_3$ should have a higher domainness score for \emph{Science}.

To the best of our knowledge, no specific measures exist to quantify this concept. In the next section, we propose several automatic metrics to measure the domainness of a collection and afterwards, in Section~\ref{sec:automatic}, we apply them to determine the quality of our in-domain collections of Wikipedia articles and we study their correlation against the manual evaluation performed in Section~\ref{sub:manual}.

\subsection{Domainness Metrics}
\label{sub:metrics}

Although there is no predefined scale to quantify the domainness, we intend to measure if a corpus represents better a domain than another one, and how or if it degrades when being enlarged. 
With that in mind, and in order to come out with a more affordable evaluation framework, we define four different families of automatic metrics inspired by the work of~\citet{Kilgarriff:01} on corpus analysis and the work of~\citet{newmanEtAl:2010} on topic coherence. The first three families intend to measure the representativity of the corpus and characterise a domain on the basis of its characteristic vocabulary. 
Quite differently, the fourth family intends to measure the cohesion of the collection 
without the requirement of characterising the domain.

\paragraph{Family 1: Density of terms.} 
We begin with the assumption that a corpus describes better a domain the higher the density of terms it contains belonging to the domain's characteristic vocabulary.
Obtaining this vocabulary is straightforward when using the Wikipedia as a corpus.
Since root articles belong to the domain by 
definition, the characteristic vocabulary can be obtained as the most frequent 
terms in this subcorpus (as it has been assumed in our models). The 
\emph{density} of these terms should be a measure of the representativity of the 
collections. 
We propose two densities based on two different term frequency estimations~\cite{Salton:1988}. The first one is the term frequency of all in-domain terms $w_i$ in the collection, $c_{\rm terms}$=$\sum\nolimits_{w_i}$${\rm counts}(w_i)$, normalised by the number of articles, $N$:

\begin{equation}
   C_{\rm terms}/N \equiv \frac{1}{N} \sum_{\rm art} c_{\rm terms}.
\end{equation}

\noindent
The second one is the augmented frequency of in-domain terms for each article normalised by the number of articles:

\begin{equation}
   \hat{c}_{\rm terms} = \frac{1}{N}\sum_{\rm art} \left( K+(1-K) \frac{c_{\rm terms}}{c_{\rm max}} \right),
\end{equation}
 where $c_{\rm max}$ are the counts for the most frequent term in each document and the optimum value of $K$ is $0$ in our experiments.

\paragraph{Family 2: Mutual Information.} 
The evaluation of the quality of a corpus regarding domainness is somehow related to the evaluation of topic models. In the first case, we have a collection of texts and we want to evaluate how well they describe a domain that might be characterised or not by a set of keywords. In the second case, we are given a set of keywords and we want to evaluate how well they describe the topic (domain) of a collection. 
\citet{newmanEtAl:2010} introduced  the concept of \emph{coherence} of a topic as the coherence or interpretability of its keywords. They measure it with the average or median of pointwise mutual information (PMI) between the topic keywords. Subsequent works use NPMI~\cite{bouma2009normalized}, a normalised version of PMI, for the same purpose:

\begin{equation}
 {\rm PMI}(w_i,w_j) = log_2\frac{p(w_i,w_j)+\epsilon}{p(w_i)\,p(w_j)+\epsilon},
\end{equation}

\begin{equation}
 {\rm NPMI}(w_i,w_j) = \frac{{\rm PMI}(w_i,w_j)}{-log_2(p(w_i,w_j)+\epsilon)}, 
\end{equation}
where $w_i$ and $w_j$ are the keywords describing a topic ---the terms in the characteristic vocabulary in our case---, $\epsilon$ is a smoothing constant, and $p$ stands for frequentist probability.
For topic modelling, the median of the pairs showed better correlation with human judgments than the mean because it is less sensitive to outliers~\cite{newmanEtAl:2010}.

We apply the two measures and two variants to evaluate domainness; assuming that the vocabulary we use perfectly describes the domain and the loss in the value of (N)PMI gives information about the background collection. We expect in-domain collections to have a high density of in-domain terms ---$p(w_i)$ and $p(w_j)$ values higher than in general collections---, but we still expect co-occurrences of terms to be representative. Computationally, the main difference with the original usage is how to estimate term co-occurrence frequencies to compute probabilities. In topic modelling, co-occurrences are sampled from the full collection or from an external source, such as the Wikipedia or Google $n$-grams, with a sliding window of length $m$ words. 
Here, we always use the full in-domain collection and consider as window an entire article of the domain: (N)PMI$_{\rm art}$. Notice that with this definition the window has a variable length. In order to study if this difference is relevant, we define a second variant (N)PMI$_{\rm col}$ where we estimate a probability as the sum of probabilities in all the articles of the collection
instead of simply the counts per article as in the original version:
\begin{equation}
 p(w_i)_{\rm art} = \frac{\sum_{\rm art} {\rm counts}(w_i)}{\sum_{\rm art} {\rm terms}}
\end{equation}

\begin{equation}
 p(w_i)_{\rm col} = \frac{\sum_{\rm art} \left( {\rm counts}(w_i)/\rm terms \right)}{N}
\end{equation}

\paragraph{Family 3: Correlations.} 
In his deep study, \citet{Kilgarriff:01} quantifies the similarity among corpora measuring 
frequencies of words and cross-entropies. Here we adapt the measure he evaluated 
as the best one fitting our problem ---the Spearman correlation--- and add  
Kendall's $\tau$ correlation as well for a better generalisation.
Spearman $\rho$ (and Kendall's $\tau$) is a non-parametric rank correlation. It measures the difference in rank order between two distributions:

\begin{equation}
 \rho = 1- {\frac {6 \sum pd_i^2}{n(n^2 - 1)}},
\end{equation}
\noindent 
where $pd$ are the pairwise distances of the ranks of the terms $w_i$ and $w_j$, and $n$ is the number of terms. For Kendall, we have:
\begin{equation}
 \tau = \frac{c-d}{\sqrt{n(n-1)/2-T}\sqrt{n(n-1)/2-U}}, 
\end{equation}
where $c$ is the number of concordant pairs, $d$ is the number of discordant pairs,
\begin{equation}
T = \sum_t t(t-1)/2 \enspace \text{and} \enspace 
U = \sum_u u(u-1)/2 ,
\end{equation}\noindent 
where $t$ is the number of times the terms $w_i$ are tied, and $u$ is the number of times the terms $w_j$ are tied.

In our particular case, we measure the difference in rank order of $n$ terms in two corpora: an 
extracted collection of articles of a given domain, and the subset of its root 
articles. Terms are defined as before; since the important feature of a term is 
its rank and not its absolute frequency, this measure can be used for corpora of varying size.

To compute the correlation, one needs to find the $n$ most frequent common terms. These are obtained as the union of the first $m$ terms for every corpus.
The terms that do not appear in the other corpus have frequency zero and are therefore ranked at the bottom of the other corpus' list. Some heuristics are considered to build the vectors:

\begin{itemize}
 \item[\Ni] At most 1000 terms from the top 10\% (if available) for every 
collection are used, therefore the maximum number of common elements is 2000;
 \item[\Nii] terms with frequency 1 are not considered within the 1000; and
 \item[\Niii] correlations are not estimated with less than 5 points.
\end{itemize}

Both Spearman and Kendall correlations measure monotonicity relationships. Although we checked that in most cases the two statistics lead to the same conclusions, 
Kendall's $\tau$ has shown to be more robust, more appropriate for small samples and, given its definition, to deal better with ties and outliers \cite{crouxEtDehon:2010}, so it is the one we use as a representative of this family. 

\paragraph{Family 4: Cohesion.}
In this case, our objective is assessing the distance between the articles pertaining to a given domain, according to our models. The lower the distance between such articles, the more cohesive they are, and the more likely that they actually belong to the domain; i.e.\ the better the model works. 
In order to come out with a single number to compare across different models, we compute the average distance between all the article pairs in the domain. Considering standard vector-space models to represent the texts could result in measures sensitive to length and vocabulary differences between the pairs of articles. Article embeddings obtained as document embeddings simply by using doc2vec~\cite{leEtAl:2014} could solve this issue, but the quality would still depend on the language because poorer languages have a lesser amount of data where to estimate the embeddings. On top of these factors, in this work we focus on multilinguality. As a result, we opt for using a high-dimensional concept-based representation, ESA.

The purpose of ESA is representing texts ---regardless of their lengths--- onto a high-dimensional concept-based space. The space is built on top of the term--document matrix $\mathbf{D}$ generated from a large collection $D$ of documents using tf-idf weighting. 
The representation of a text is then built by comparing it against $\mathbf{D}$, resulting in a $|D|$-dimensional vector. For efficiency reasons, the average distance is computed with respect to the center of the collection as
\begin{equation}
 d_{\rm ESA} = \frac{1}{N}\sum_{a} dist_\theta(a_{ESA}, c_{ESA}) ,
\end{equation}
where $a_{ESA}$ is the vector representing article $a$ and $c_{ESA}$ is the centroid of all the vectors in the corpus and $dist_\theta$ refers to the angular distance:
\begin{equation}
 dist_\theta = \arccos \left( \frac{a_{ESA} \ldotp c_{ESA}}{ \parallel a_{ESA}\parallel \parallel c_{ESA}\parallel} \right).
\end{equation}

\section{\textbf{\textit{Domainness}} Evaluation}
\label{sec:automatic}

\begin{figure*}[t]
 \centering
 \includegraphics[width=1.00\textwidth]{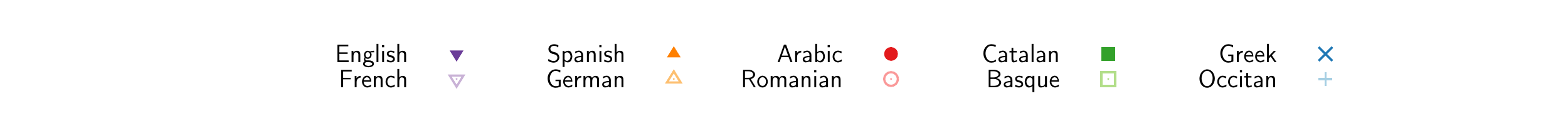} \\[-5mm]
 \includegraphics[width=0.46\textwidth]{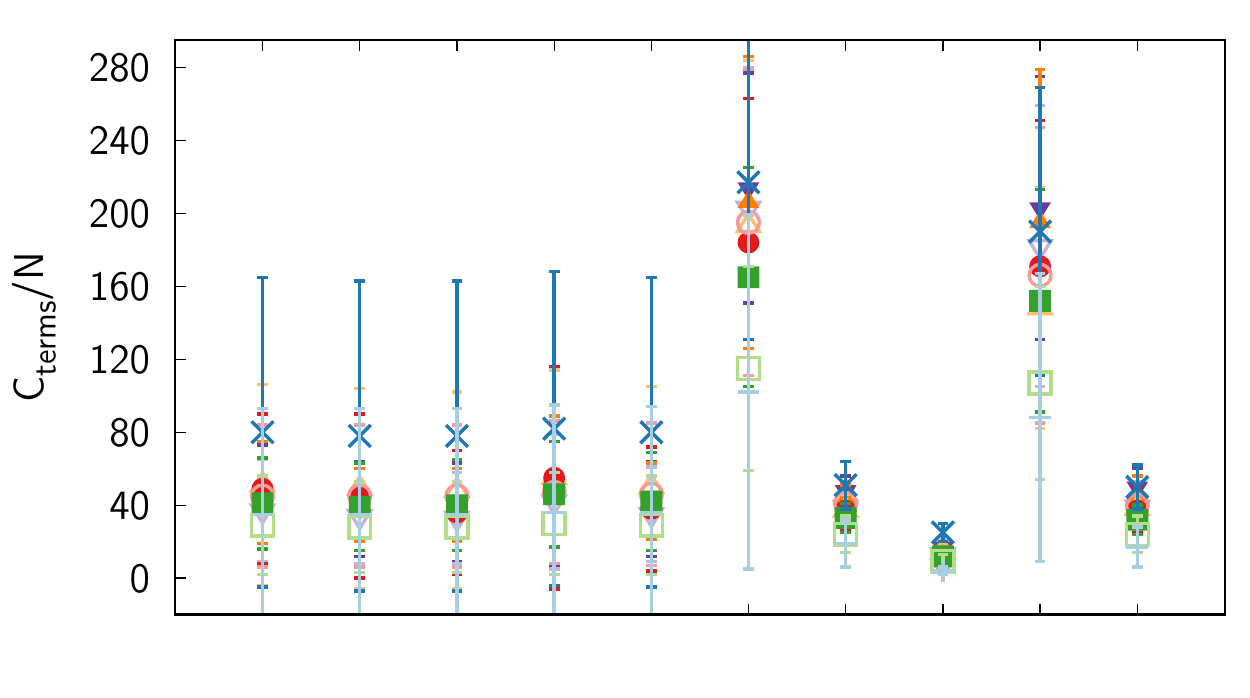} \hspace{2em}
 \includegraphics[width=0.46\textwidth]{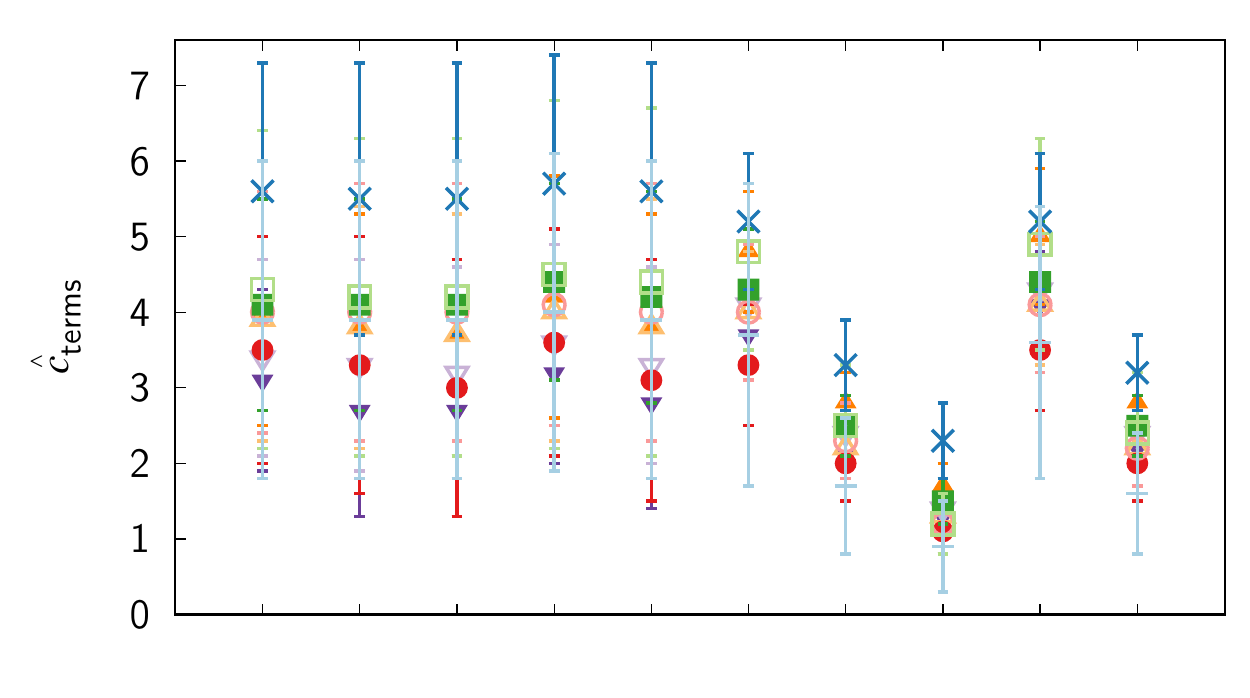} \\[-5mm]
 \includegraphics[width=0.46\textwidth]{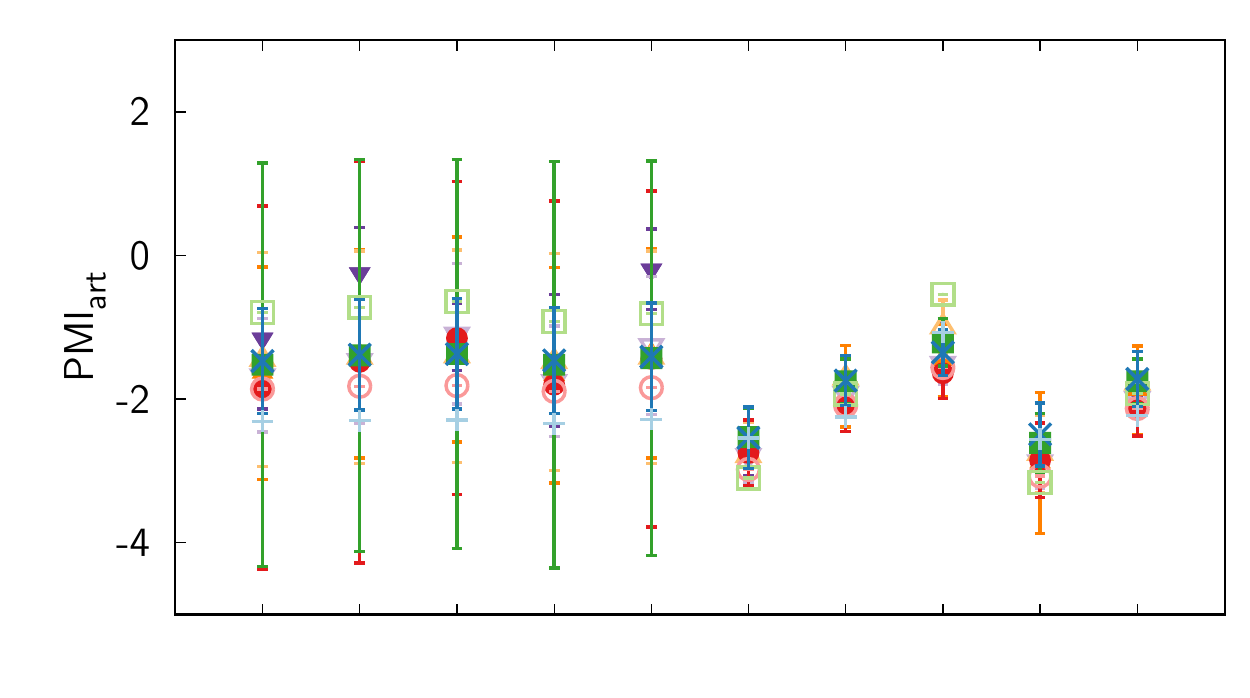} \hspace{2em}
 \includegraphics[width=0.46\textwidth]{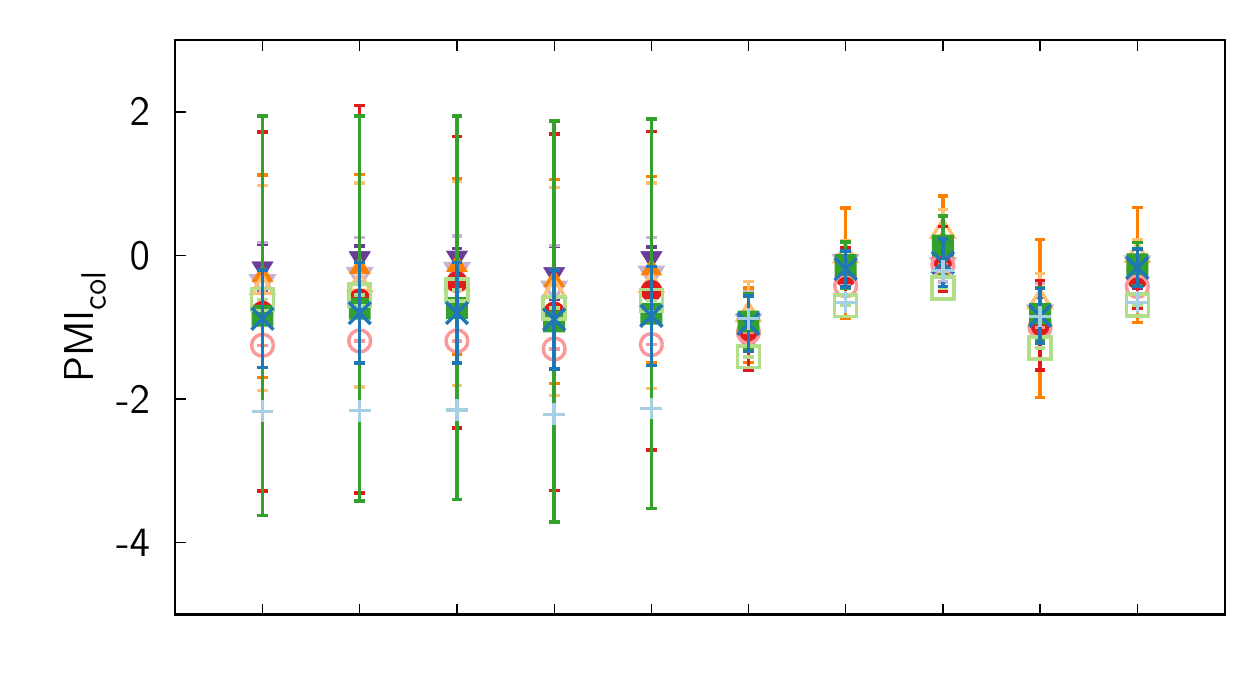} \\[-5mm]
 \includegraphics[width=0.46\textwidth]{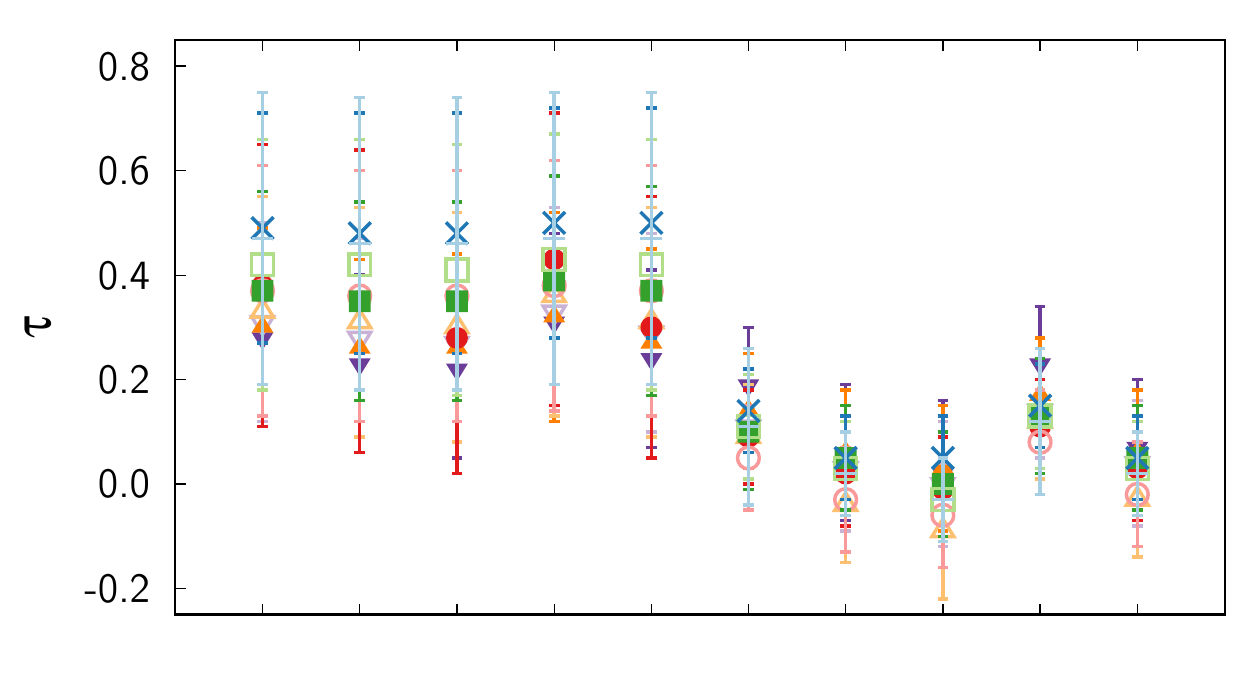} \hspace{2em}
 \includegraphics[width=0.46\textwidth]{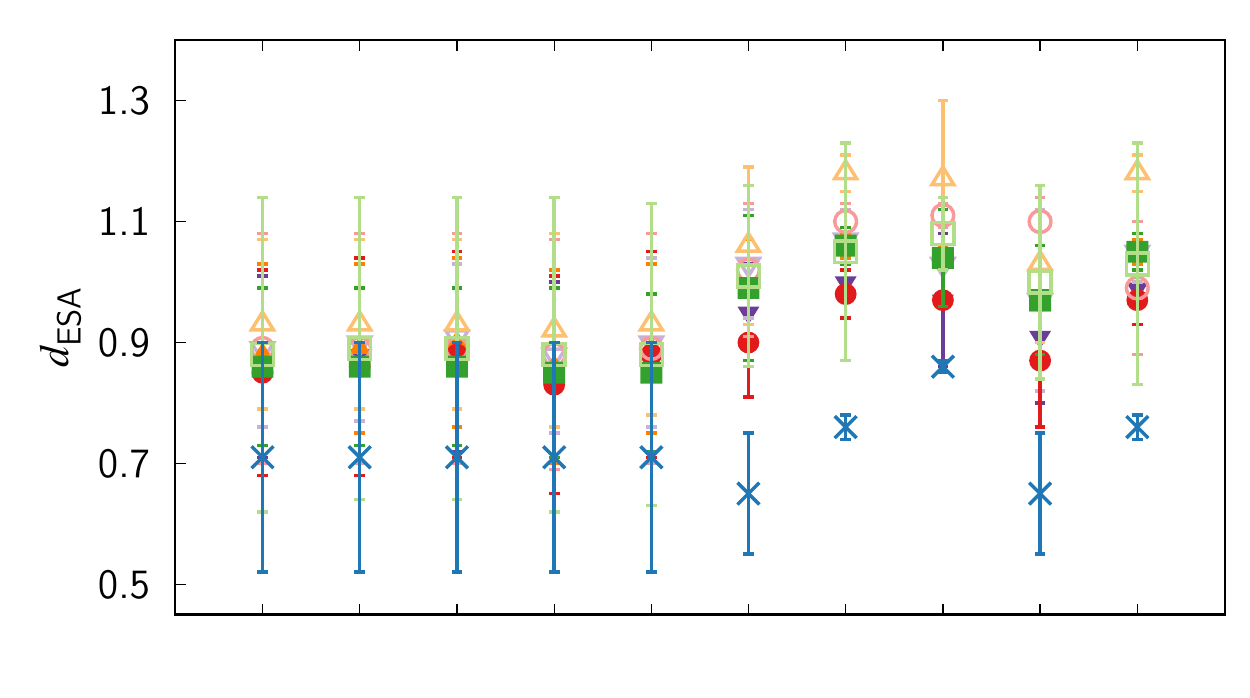}\\[-3mm]
 \vspace{-0.8em}
 \includegraphics[width=0.46\textwidth]{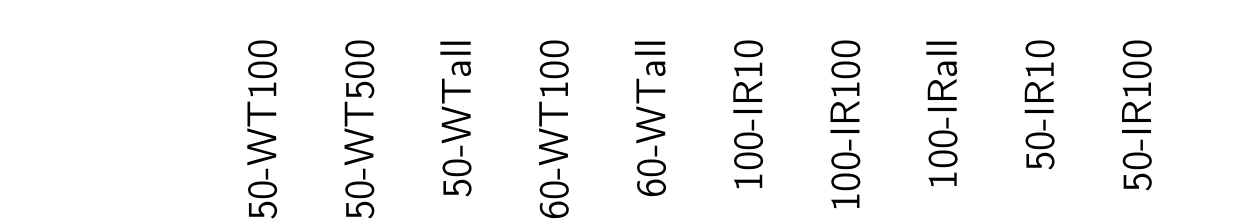} \hspace{2em}
 \includegraphics[width=0.46\textwidth]{models_metric.pdf}
 \bigskip
 \caption{
 Automatic evaluation of the in-domain collections for the different systems and languages under study with six measures which are representative of the four families introduced in Section~\ref{sec:evaluation}. Points represent the arithmetic mean over the 743 selected domains.}
\label{fig:metrics}
\end{figure*}

Now we inspect the numbers obtained for the different metrics when analysing the collections extracted by the WP and IR models in all languages and domains.
Figure~\ref{fig:metrics} summarises the results with some representative measures of the four families of metrics. We plot the mean and standard deviation of six measures, $C_{\rm terms}/N$, $\hat{c}_{\rm terms}$, PMI$_{\rm art}$, PMI$_{\rm col}$, $\tau$, and \clesa, for the ten languages under study and the 10 systems analysed. 
For comparison purposes, we also chose a representative model of every family (\wtV{100}\ and \irC{10}) and compare it against a subcollection of the other family gathered to have the same size. Although we do not include the corresponding figures, the outcomes are also discussed.

For the representativity measures (Families 1, 2 and 3), the size of the characteristic vocabulary used in the experiments is 100 terms, i.e.\ 5049 term pairs. In all cases, the collections on which probabilities are estimated are preprocessed as explained in Section~\ref{sub:graph-selection} so that the format of the articles matches the terms.

\paragraph{Family 1. }

By design, IR systems are the ones with a larger number of in-domain terms. The density is expected to be higher in the smallest collections, \irall{10}, because they contain the top ranked articles retrieved according to these terms. In WT systems, the terms have a high density in the root articles, also by definition of the model, but there is no expectation for a high number of in-domain terms in the rest of the collection. The output of $\hat{c}_{\rm terms}$ and especially of $C_{\rm terms}/N$ reflects this (cf.\ top-left plot in Figure~\ref{fig:metrics}).
Differences between WT systems seem not to be significant under these metrics. In general, differences appear for large editions, where the vocabulary size varies notably from system to system. 
The best WT system is \wtVI{100}, the most restrictive one and that with less articles per collection, with a mean across languages of  $C_{\rm terms}/N=49.7$ and $\hat{c}_{\rm terms}=4.1$.
However, \wtVI{all}\ has a higher density of in-domain terms than any of the 50-* systems for some editions (those with less categories) even if the obtained corpora are larger.

As expected from its definition, IR systems with the smallest collections (\irall{10}) are clearly the best ones according to $C_{\rm terms}/N$, the normalisation in  $\hat{c}_{\rm terms}$ smooths the effect and makes systems closer to each other. 
Since IR collections grow significantly after allowing for lower retrieval scores, there are a lot of differences between IR models. According to these metrics, \irall{10} systems are better in \emph{quality} than any WT model, especially for large editions, with the additional benefit that they gather larger collections. This effect is more pronounced when comparing equal-size collections, but disappears for the less constrained configurations where WT models are better. 
If we analyse the results per edition, Greek is the language on which both models perform the best. There is no clear trend for the other editions, although English and Arabic perform poorly in contrast with the others. This is one of the differences when evaluating with the correlation family of metrics (family 3). In this case, English, Greek and Spanish are the editions for which results are the best. This is a first indication that both metrics are not equally valid for assessing the quality of the extractions.

\paragraph{Family 2.}
Contrary to in-domain terms, there is no requirement on the number of co-occurrences of terms when building the systems, neither for IR nor for WT systems. 
The plots in the middle row of Figure~\ref{fig:metrics} show the mean and standard deviation of PMI$_{\rm art}$ and PMI$_{\rm col}$.
One would expect positive PMIs for related terms, meaning that they occur more frequently together than if they were independent in a general collection, but we obtain negative values for most collections. The reason is the high density of in-domain terms in all the documents, which causes co-occurrences to have comparatively less weight than in general collections. 

Since we want to indirectly evaluate the collection and not the terms, we just compare the values of the different models. Within a family of systems, WT or IR, the scores completely depend on the size of the collection, the larger the collection the better the evaluation. When comparing the two systems, WT systems are better than IR systems even if IR collections tend to be larger. For instance  PMI$_{\rm art}$=-1.1$\pm$1.0 for the \wtV{100}\ English collection, with a mean of 50,514 documents per domain, and PMI$_{\rm art}$=-2.8$\pm$0.3 for \irC{10} with a mean of 64,239 documents per domain. The values of PMI$_{\rm col}$ for these collections are -0.2$\pm$0.3 and -1.2$\pm$0.4. We observe the same trends with PMI$_{\rm art}$ and PMI$_{\rm col}$, but the scores with PMI$_{\rm col}$ tend to be higher. When we estimate the normalised PMIs, differences among models become smaller, but the main conclusions hold.

If we look at differences across languages, we see that the scores are almost independent of the language for IR systems, whereas English collections are the best ones for WT systems and the Romanian and Occitan the worst ones.
Besides, Romanian, Basque and Occitan have large deviations, especially in WT systems. In IR systems, these languages have the smallest collections, but this is not the case of WT\@. The uncertainties for these languages which range from $\pm4$ to $\pm8$ are not shown in Figure~\ref{fig:metrics} for clarity.

\paragraph{Family 3.}
As observed in the bottom-left plot of Figure~\ref{fig:metrics}, correlation measures show a clear preference for the WikiTailor model. Kendall's $\tau$ lies in the range $[0.2, 0.5]$ for WT systems and $[-0.1,0.2]$ for IR systems. Results are equivalent with Spearman's $\rho$ although with a higher score: within $[0.3, 0.6]$ for WT systems, $[-0.1, 0.3]$ for IR systems. 
For different variations of a model, the results are consistent with those seen with the measures related to the density of terms: smaller and more constrained collections are always evaluated better. However, the standard deviation is too big to make statistically significant statements when comparing models within one same family.
In general, the \emph{quality} increases for Wikipedia editions that have less categories 
for WT systems; whereas there is no specific trend for IR systems.
Large editions correlate less because their domains have more articles; when only domains with more than 100 articles are considered, correlations diminish for those languages where this is important, such as Occitan, Greek, or Basque; and the scores per language become more homogeneous.
When we compare IR and WT collections up to an equal size, we confirm that WT models are better than the IR ones according to $\rho$ and $\tau$ and, the smaller the edition, the more evident the difference becomes.

\paragraph{Family 4.}
Following the original ESA proposal and in consistency with this work, we use 
the Wikipedia as our reference text collection $D$ for the cohesion-oriented 
metric. The size of $D$ for each of the languages under study is $12,539$, as 
this is the size of the intersection among the top nine Wikipedia language 
editions. 
\citet{GottronEtAl:2011} showed the convergence of the method with $10,000$ articles approximately, so, we discard the tenth edition, the Wikipedia in Occitan, because including it would decrease too much the number of articles in $D$. The Occitan models are therefore not evaluated using this measure.

Similar trends seen with the previous metrics regarding quality can be observed with \clesa, even if its nature its different. In this case, lower values imply collections with a higher cohesion, irrespective of the domain they belong to. The results are shown in the bottom-right plot of Figure~\ref{fig:metrics}. Since WT collections include the root articles of the desired domain and IR systems retrieve only articles that contain the vocabulary of the domain, we can assume that a large cohesion implies a large domainness.  As it happens with $\rho$ and $\tau$, \clesa\ clearly peaks WT models (\clesa$\approx$0.85) over IR ones (\clesa$\approx$1.00). The best (worst) collections are obtained for Greek (German). Again, mean averages do not allow to establish preferences among the different configurations within a same family of models in a statistically significant way, but models with the smallest set of terms (\irall{10}\ and \wtall{100}) are preferred; i.e.\ more constrained collections have a larger cohesion. 

\medskip
All the metrics we have defined clearly differentiate the quality of WikiTailor and IR systems when we study the average on all the domains but only show trends within the different models of a same family. In general, the most constrained configuration for each family (\wtVI{100}\ and \irV{10}) obtains the most in-domain collection, but the difference is sometimes minimal with respect to another configuration which, on the other hand, might have retrieved many more articles. We are comparing 7,430 collections for 10 different models but, in practice, a standard user will be dealing with only a few of them. In that case, it might be more fruitful to decide which is the collection to be used according to the scores but also according to size and domain representativity requirements.
Notice also that the density metrics (families 1 and 2) behave differently to correlation (family 3) and cohesion (family 4) measures when dealing with the most constrained collections.

\bigskip 

The human judgments from Section~\ref{sub:manual} also allow us to estimate the quality of the automatic evaluation metrics. We calculate the Pearson 
correlation $r_P$  between the crowdsourced precisions and the scores given by the automatic metrics on the same subcollections  considering 200 articles per system and language in three domains (settings in Section~\ref{sub:manual}).

\begin{figure*}[t]
 \centering
 {\scriptsize \hspace{1cm}$\color{orangeplot}{\blacklozenge}$ \irC{10}\ ~~~~~~ $ \color{greenplot}{\bullet}$ \wtV{100}}\\
 \smallskip
 \includegraphics[width=0.44\textwidth]{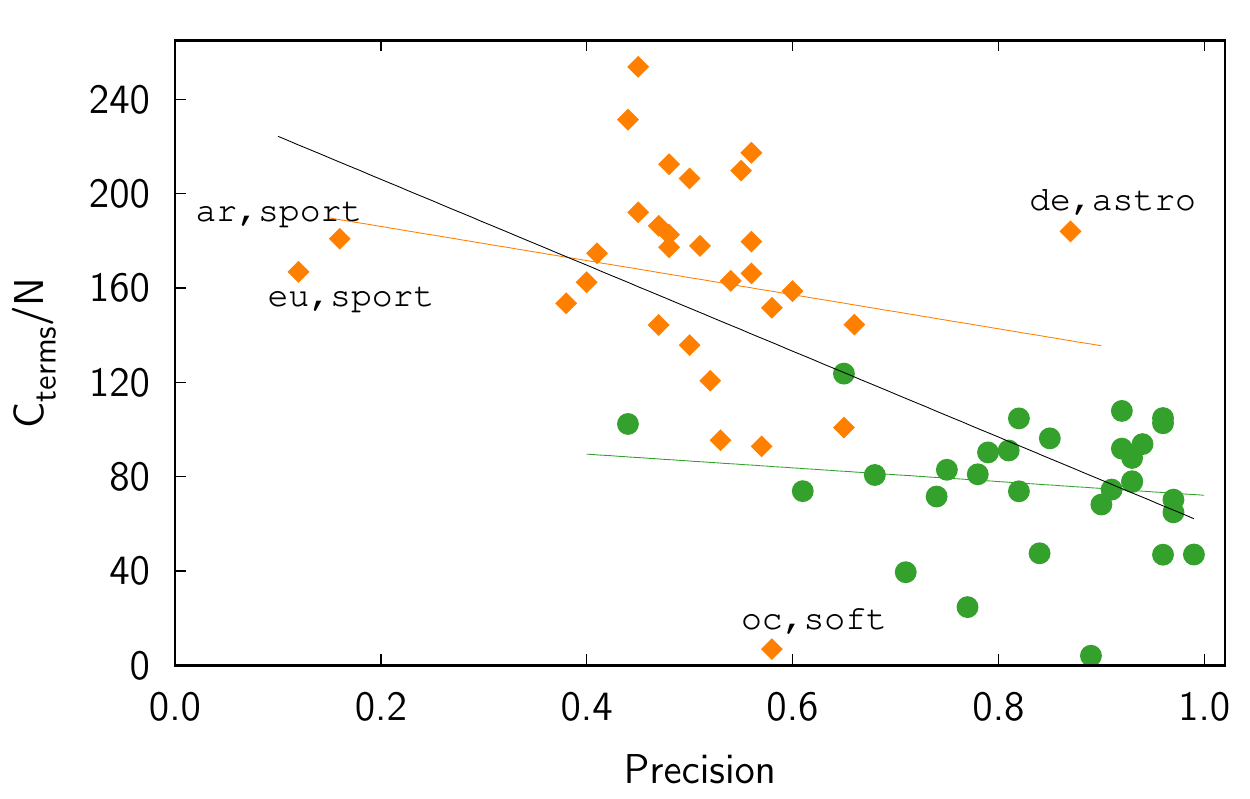}\hspace{2em}
 \includegraphics[width=0.44\textwidth]{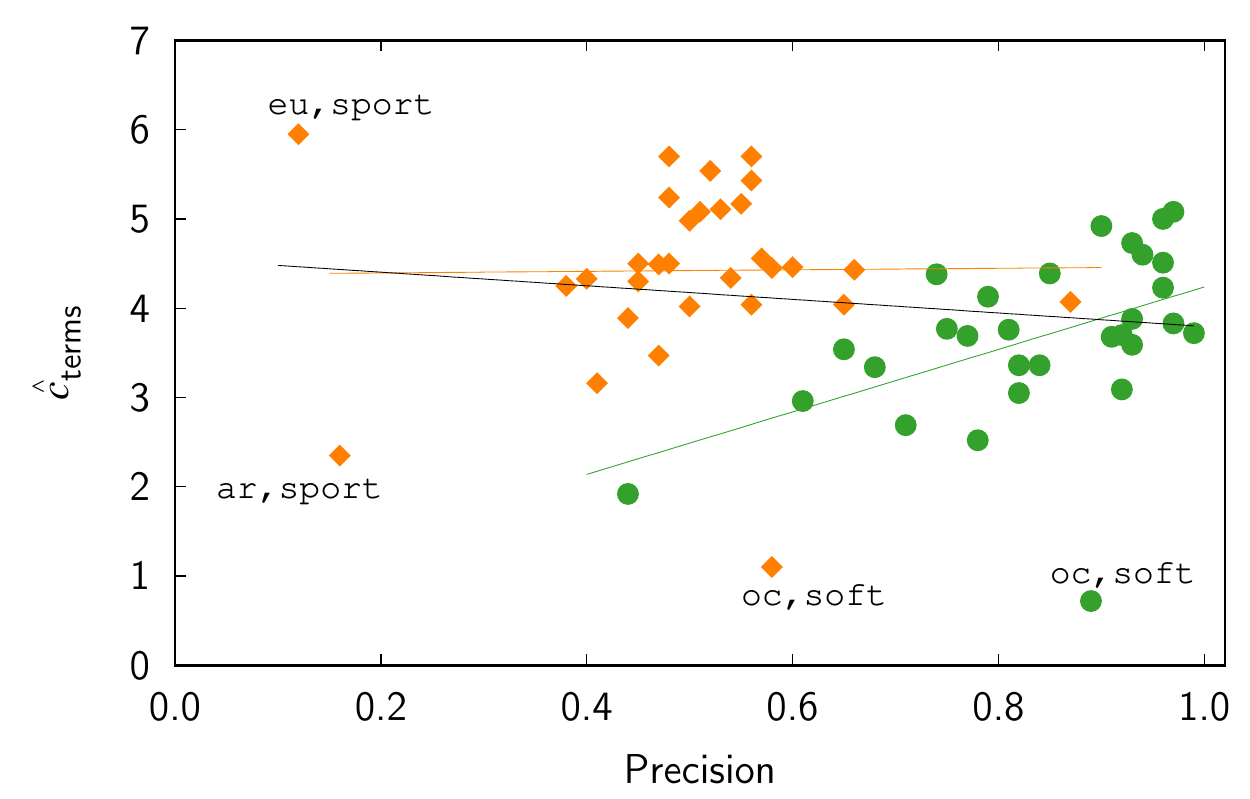} \\
 \includegraphics[width=0.44\textwidth]{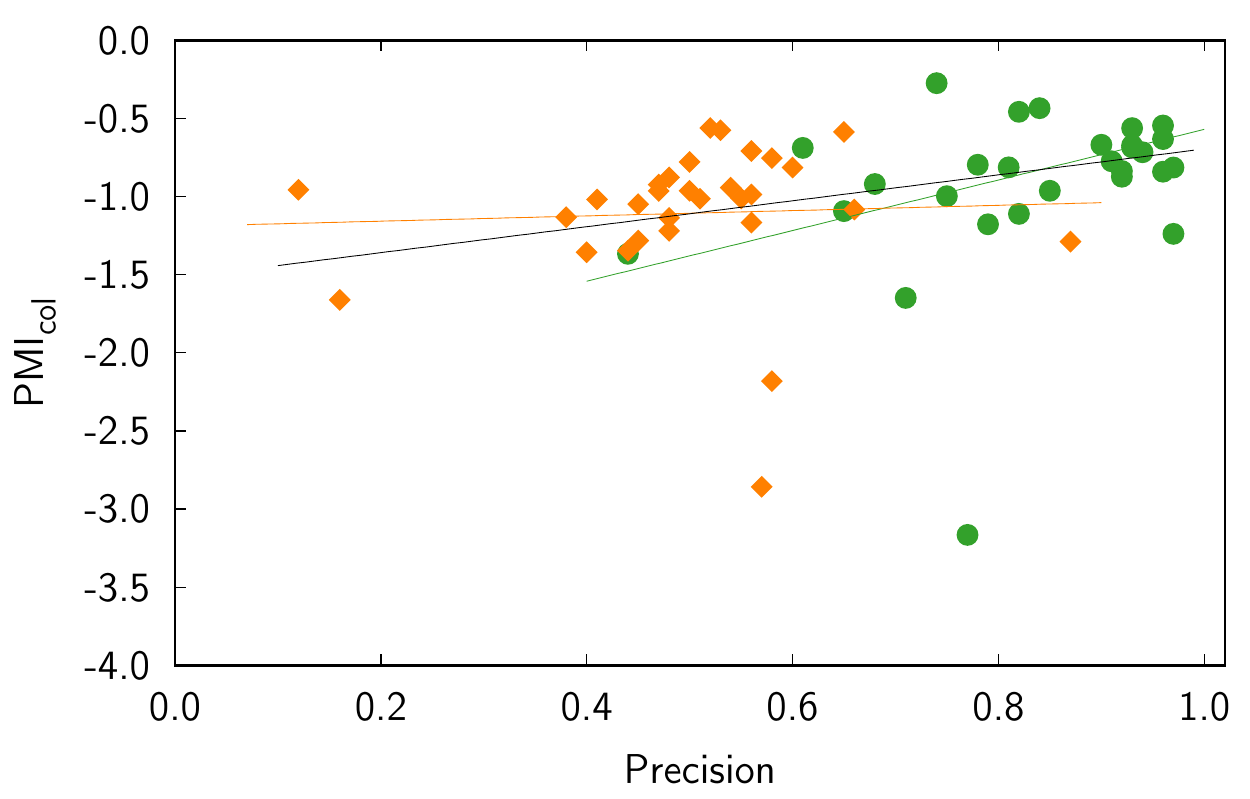}\hspace{2em}
 \includegraphics[width=0.44\textwidth]{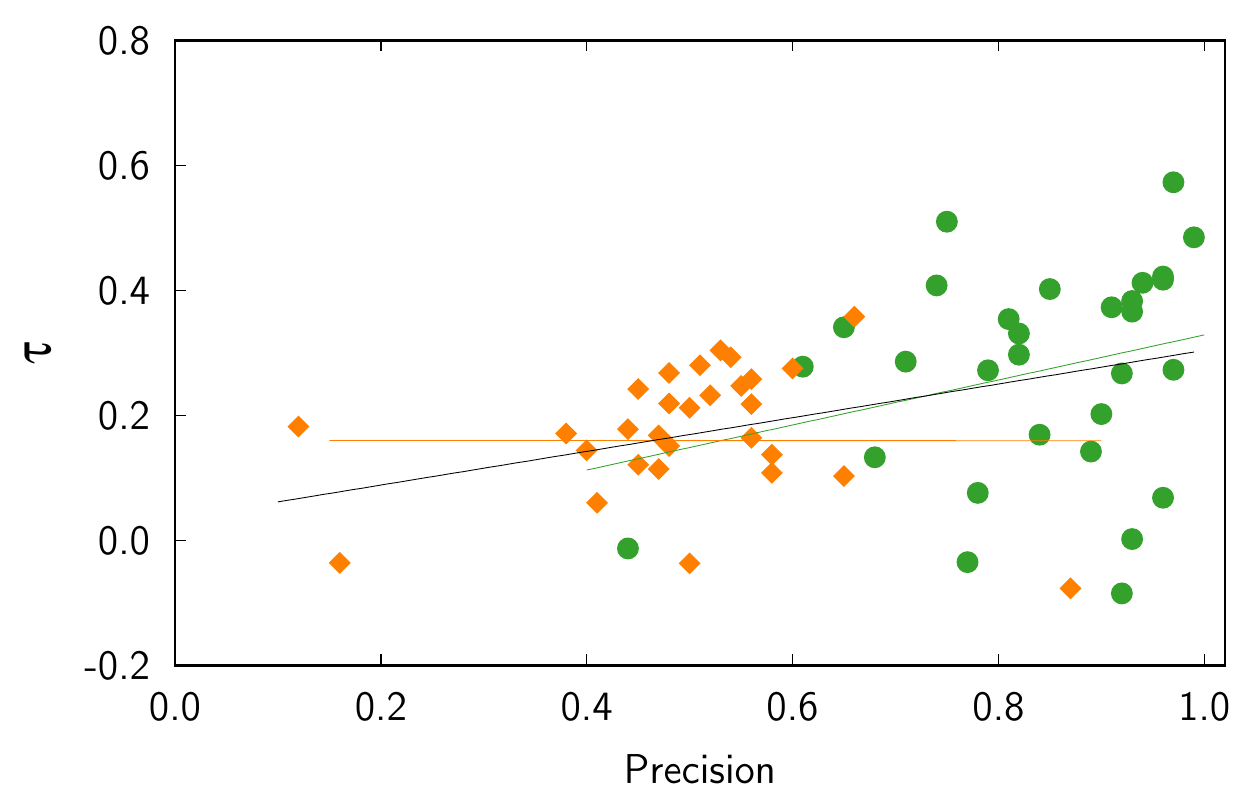} \\
 \includegraphics[width=0.44\textwidth]{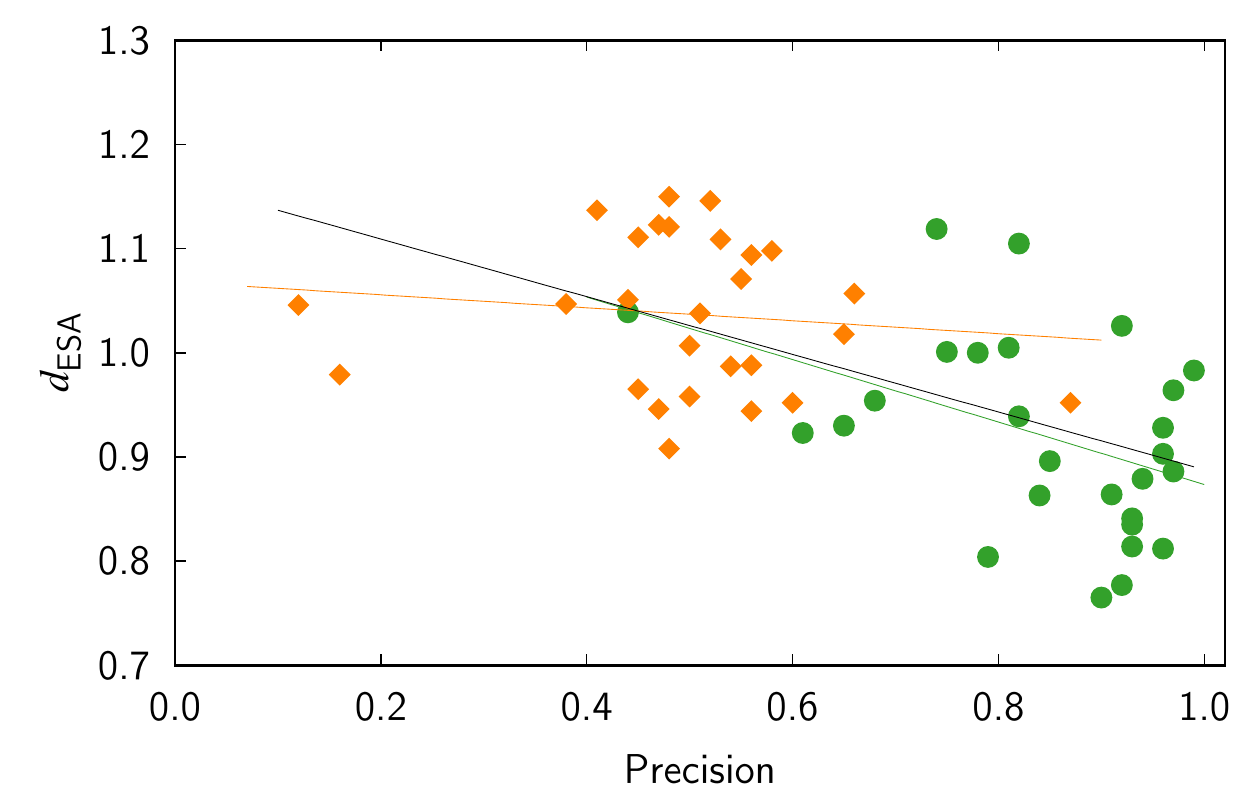}\hspace{2em}
 \includegraphics[width=0.44\textwidth]{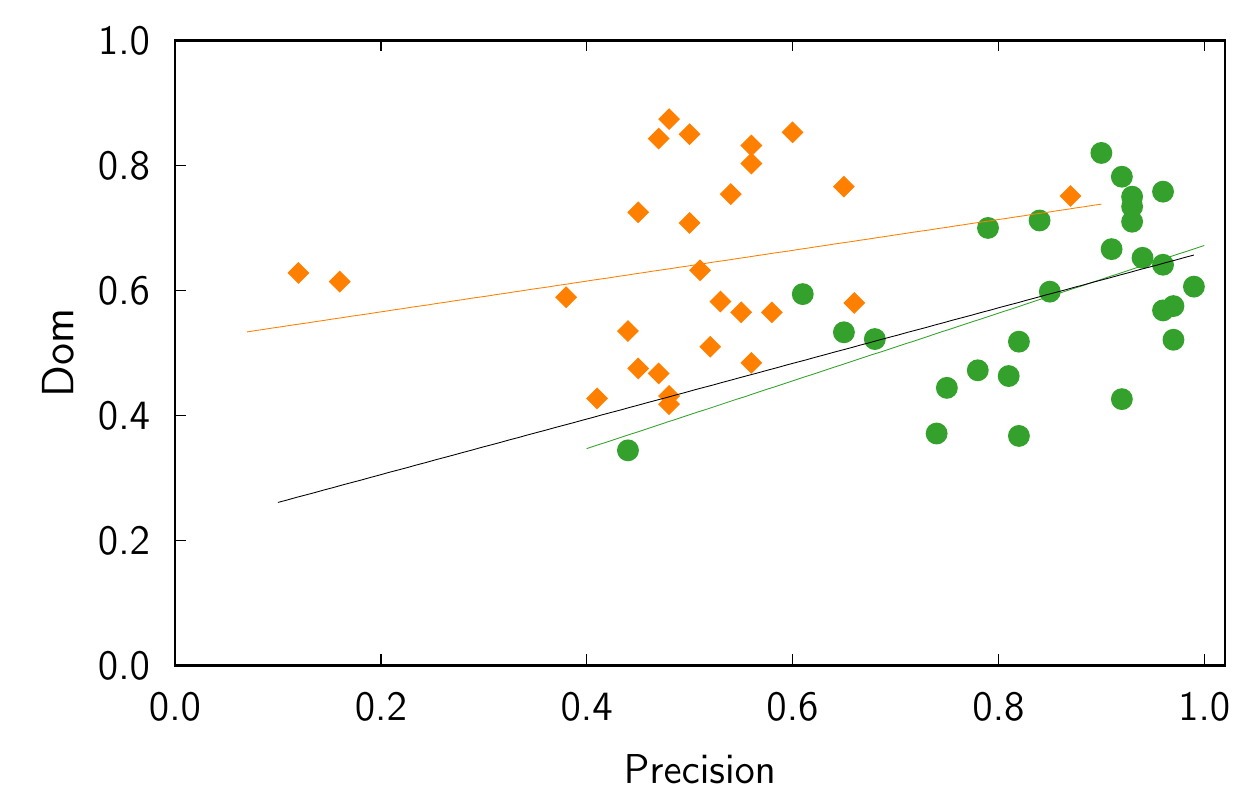}
 \medskip
 \caption{Relation between six domainness measures and the precision given by human judgments (see text for correlations). Points correspond to the score for the 10 languages in the three domains manually evaluated, some examples are highlighted.}
\label{fig:correlations}
\end{figure*}

A visual inspection of the data is a first good clue to understand the behaviour of the metrics.
Figure~\ref{fig:correlations} shows the relation against soft precision of six metrics: $C_{\rm terms}/N$, $\hat{c}_{\rm terms}$, PMI$_{\rm col}$, $\tau$, \clesa, and a full measure for domainness: \emph{Dom}. 
The first worth-noticing aspect is that in all cases the graphical counterpart of Table~\ref{tab:cf-precision} (e.g., points corresponding to the 50-WT100 system; green bullets) are located towards higher precision values than those corresponding to the 100-IR10 system (orange diamonds). We plot 60 points per figure, corresponding to two systems applied on ten languages $\times$ three domains. The exceptions are \clesa and \emph{Dom}, for which only nine languages $\times$ 3 domains are shown (we discard those collections with less than 200 articles for the correlation estimation (\emph{Astronomy} and \emph{Software} for Occitan, and \emph{Sport} for Basque; cf.\ Table~\ref{tab:cf-precision}).

\paragraph{Family 1.}
Counterintuitively, the metric with the highest and negative correlation is the density of terms $C_{\rm terms}/N$ with $r_P=-0.716$. The high value is just an artifact given by the different composition of the WT and IR collections.
By construction, the IR system retrieves articles with lots of terms, whereas the dependence for WT models is lower. The quality of WT is better, so there is a clear anticorrelation between the density of terms and the precision. If we look at what happens only within WT or IR instances (i.e.\ only with the green or orange points independently), we obtain worse correlation values: $r_P=-0.18$ for WT and $r_P=-0.23$; still negative in both cases, but closer to zero. The fact that these values are not positive invalidate the assumption we made to use this family of metrics to measure domainness. The results show how the density of the characteristic vocabulary of the domain is neither a sufficient nor a necessary condition to obtain in-domain corpora. It can be a good estimator for the representativity of the corpus, but if the cohesion is low, the domainness is also low.

The additional normalisation of this measure included in the augmented frequency  $\hat{c}_{\rm terms}$ rules out the metric as a global measure. The Pearson correlation for $\hat{c}_{\rm terms}$ when all the data are used together is $r_P=-0.08$: these two variables do not correlate. Since the frequency of terms is now normalised to the most frequent term, their importance is lower,
and therefore, both WT and IR behave similarly, with slightly higher values for IR than for WP\@. The reason is the same as before, hence exhibiting an anticorrelation with precision scores.
However, when looking into the two systems, the correlation increases specially for WT:  $r_P=0.63$ for WT and $r_P=0.36$ for IR\@. So, within a system, we have a positive correlation of  $\hat{c}_{\rm terms}$ vs \emph{Precision} which indicates that  $\hat{c}_{\rm terms}$ is a good barometer of the quality of a WT extracted in-domain corpus.

\paragraph{Family 2.}
Metrics related to mutual information or co-occurrence show a clear positive trend with respect to precision. 
Even with negative PMI values, human judgments show how the best collections have higher PMI values. The score that correlates best with precision is PMI$_{\rm col}$ with $r_P=0.57$. The metric with the standard probability calculation PMI$_{\rm art}$ is close with $r_P=0.55$. The variable size sliding window that we use, an article, is not affecting the results. The normalised versions are slightly below these values because the effect of the normalisation is to smooth differences among points (NPMI$_{\rm art}$ has $r_P=0.41$; NPMI$_{\rm col}$ has $r_P=0.55$). 
We also observe that in our setting the median of (N)PMI is a better estimator than the average. 

If we compare the subset of points belonging to WT and IR, the correlation is lower than the global one in both cases but specially for IR, where we observe no correlation between the metric and the observations (PMI$_{\rm art}^{\rm WT}$ has $r_P=0.44$; PMI$_{\rm art}^{\rm IR}$ has $r_P=0.08$). Notice that the different nature of WT and IR allows us to say that a high density of in-domain terms in an article does not imply that it belongs to the domain, as concluded from the fact that $C_{\rm terms}/N$ and $\hat{c}_{\rm terms}$ for the IR system are above their equivalents for WT\@. However, a higher number of co-occurrences of the domain vocabulary does (PMI$^{\rm WT}$ larger than PMI$^{\rm IR}$).

\paragraph{Family 3.}
The next family of metrics, $\rho$ and $\tau$, measures the rank correlation between the terms of an extracted in-domain collection and a collection of Wikipedia root articles in the same domain. The correlation with soft precision is in this case $r_P=0.31$ for $\rho$, and $r_P=0.34$ for $\tau$. As the plot for $\tau$ in Figure~\ref{fig:correlations} shows, the dispersion of the WT points is larger, but their subset has a higher correlation than the IR one ($r_P=0.25$ vs $r_P=0.02$). For the IR subset, the metric is a very bad measure of the quality of the extraction, but contrary to the augmented term frequency metric  $\hat{c}_{\rm terms}$, it performs better in the global setting than within the subsets.

\paragraph{Family 4.}
The measure of cohesion of the corpus through ESA distances results in a good estimator. With a global correlation of  $r_P=-0.60$ and subset correlations of $r_P=-0.41$ (WT) and $r_P=-0.13$ (IR), \clesa\ is the best individual metric to estimate the domainness of a collection in general, but  $\hat{c}_{\rm terms}$ is the best metric when we focus on WikiTailor extractions. $\hat{c}_{\rm terms}$ is not bounded. Its range is [0, $\infty$), where high densities imply a good quality. However, due to the lack of top boundary, it is useful to compare collections, but no clear interpretation exists in terms of an absolute number. In terms of ease of use, both \clesa\ and $\hat{c}_{\rm terms}$ rely on the Wikipedia. $\hat{c}_{\rm terms}$ comes for free with a WT extraction because we estimate the characteristic vocabulary in our models. \clesa\ performs better globally, but the cost is the need to define a reference collection, which can be different across languages.  PMI$_{\rm col}$ alleviates this problem being also language independent, but the quality its a metric is slightly lower.

\medskip
Finally, we estimate the domainness as the combination of the most promising metrics for representativity and cohesion:
\begin{equation}
 {\rm domainness} \equiv {\rm Dom} = \left( {\rm \widehat{PMI}}_{\rm col} +  \widehat{d}_{\rm ESA} \right)/2,
\end{equation}

\noindent
where hats in $\widehat{\rm PMI}_{\rm col}$ and $\widehat{d}_{\rm ESA}$ represent a normalisation of the data points between [0,+1].
As expected, we obtain the largest global correlation with the combination as representativity and cohesion are two perpendicular features. 
{\rm Dom} reaches a correlation of $r_P=0.71$ when all 60 datapoints are used. At system level, with two sets of 30 datapoints, ${\rm Dom}^{\rm WT}$ has $r_P=0.55$ and ${\rm Dom}^{\rm IR}$ $r_P=0.27$ showing that the more homogeneous a collection of points is, the less important is the combination of aspects. In that case, the correlation is slightly worse than that given by the simple augmented term frequency metric $\hat{c}_{\rm terms}$ as seen before.

\section{Summary and Conclusions}
\label{sec:conclusions}

Several multilingual applications benefit from in-domain corpora, but gathering them usually requires a considerable amount of work. We therefore design a system to extract such corpora from the Wikipedia, a multilingual online encyclopaedia with information of the domain of the articles encoded in their category tags.
The WikiTailor system explores Wikipedia's category graph and performs a breadth-first search departing from the category associated to the desired domain. From this point, it extracts all the articles belonging to its children categories down to an estimated optimal depth.
We compared the performance of WikiTailor with a standard IR system based on 
querying the Wikipedia with a set of keywords that describe the domain.
The keywords or in-domain vocabulary were extracted in the same way for the two architectures as the most frequent terms in the root articles; that is, the articles belonging to the top category.
The two methods are very different in nature and generate complementary collections. WT collections, which are smaller, are not in general a subset of the IR ones.
The experimental analysis on 10 languages and 743 domains showed the preference by automatic and manual evaluations for the WT models with respect to the IR ones. 

The manual evaluation was carried out on three domains ---\emph{Astronomy}, \emph{Software}, and \emph{Sport}--- on one model for WT and one for IR.	
Turkers in Figure Eight were asked to indicate if an article belonged to the domain or not, for a total of 200 articles per language and system. Precision was afterwards used to evaluate the quality of each collection. With an average precision of P$^{\rm WT}$=0.84$\pm$0.13 and P$^{\rm IR}$=0.50$\pm$0.14, WikiTailor resulted statistically better than the IR system. 

The lack of metrics to measure the domainness of a corpus caused an automatic evaluation more complicated. 
Therefore, we first define the concept  as a combination of the representativity and coherence of the texts in a corpus and, afterwards, we introduce several metrics to account for it.
Representativity is measured on the basis of the characteristic vocabulary of its intended domain (density, co-occurrence or correlations between distribution of terms) and coherence on the basis of the distance between the articles of the collection.
Via the correlation with human judgments, we show how the density of the characteristic vocabulary of the domain is neither a sufficient nor necessary condition for in-domain corpora. IR systems, with a higher density of in-domain terms by construction, are worse for all languages and domains in our manual evaluation. 
On the other hand, distances between the documents of a collection as measured by ESA representations outperform term-based measures and show a moderate correlation with observations.

Mathematically, we introduce Dom, a metric which is a normalised linear combination between the best representativity metric ($\widehat{\rm PMI}_{\rm col}$) and the distance-based one for coherence ($\widehat{d}_{\rm ESA}$). This combination shows a strong correlation with human evaluations, 0.71.
In summary, \clesa\ is the best individual metric to estimate the quality of a collection in general, when comparing heterogeneous collections as different in nature as the ones we explore. However, it is only measuring the coherence between the documents and the performance is improved when combined with a measure of the importance of in-domain term co-occurrences.
Within a system conclusions change. WT systems extract the articles without any request on the number of in-domain terms that the documents have, and within these collections the occurrences and co-occurrences of terms are relevant. For homogeneous collections (WT \emph{or} IR) $\hat{c}_{\rm terms}$ is the best metric.
For heterogeneous collections (WT \emph{and} IR) \clesa\ and Dom are the best options, 
meaning that coherence is more important when discrepancies in the number of in-domain vocabulary are not huge.

All the metrics and the WT and IR systems are freely available in the~\wt package.

\bibliographystyle{acl_natbib_nourl}
\bibliography{biblioWT}

\appendix

\section{Wikipedia-Specific Concepts}
\label{app:concepts}

\begin{description}
 \setlength\itemsep{-0.2em}
 \item[Category] Tag present in a set of articles grouped together by covering similar topics. 
 \item[Dump] Snapshot of an edition in the form of wikitext source and metadata embedded in XML.
 \item[Edition] Each one of the Wikipedias for a specific language.
 \item[Inter-language link/langlink] A link in a Wikipedia article towards an 
 equivalent entry in a different language.
 \item[Main namespace] The namespace in the Wikipedia containing the actual contents: the articles. Other namespaces are user, help, or category.
 \item[WCG] Wikipedia category graph. Directed acyclic graph formed by the category tags.
\end{description}

\section{Crowdsourcing Settings}
\label{app:crowdflower}

Setting up the Figure Eight crowdsourcing annotation involves four steps: \Ni the selection of Turkers, \Nii their instruction, \Niii setting the task itself and \Niv a quality control of the annotation.

\begin{table}[t]
 \caption{Geographical settings for the Figure Eight workers selection (when a language-based filtering was not available).}
 \label{tab:cf-geo}
 \medskip
 \small
 \begin{tabular}{lp{53mm}}
  \toprule 
  \bf Language & \bf Regions \\
  Spanish	& Spain, Portugal, Latin America, France \\    
  Romanian	& Romania \\    
  Catalan	& Andorra, Spain, Portugal, Latin America, France\\    
  Basque	& France and Spain \\    
  Greek		& Albania, Bulgaria, Cyprus, Greece, Macedonia, Turkey, Germany, United States of America\\    
  Occitan	& France \\
  \bottomrule
 \end{tabular}
\end{table}

The selection of the Turkers was made by their language knowledge.
We opted for three different criteria based upon language capabilities or region to determine the population that annotating each language. No language or geographical limitation was set for English, composing our most flexible 
configuration. For Arabic, French, and German we selected the corresponding 
language on the platform interface. Such a setting was not available for the 
rest of languages%
\footnote{Spanish is an exception. In that case, we opted for shaping the 
demographics geographically, as speakers of Romance languages can often 
read contents in another Romance language.}; hence we opted for a geographical 
configuration. Table~\ref{tab:cf-geo} summarises the geographical configurations, set according to four criteria: countries where the language is official (e.g., Spain for Spanish), countries with official languages from the same family (e.g., France for Catalan), neighbouring countries (e.g., Bulgaria for Greek), and countries with a high rate of immigration of native speakers (e.g., Germany for Greek).

We set the job as a binary classification task where Turkers had to assess if a Wikipedia article matches the domain displayed in the interface or not.

\medskip
  Instruction:\\
  \indent
 \framebox[.95\columnwidth][l]{%
  \begin{minipage}{.90\columnwidth}
    {\bf Task}	\\
    {- Identify the category a given Wikipedia article belongs to. 
    It either belongs to domain $d$ or to \emph{other}, where $d$ can be \emph{Astronomy}, \emph{Software}, or \emph{Sport}.}
    \end{minipage}
     }
\medskip

\noindent
The Turkers had to scroll an actual Wikipedia article, which we framed into the interface, to judge. 

After a pilot experiment, we wrote additional specific guidelines for each of the three domains aiming at clarifying 
how some ambiguous cases
should be handled by the annotators:

\medskip
\framebox[.95\columnwidth][l]{%
\begin{minipage}{.90\columnwidth}
{\bf Astronomy}	\\
- The biography of an astronomer should be considered within the Astronomy domain.\\
- Articles about Physics should not always be considered as Astronomy even if 
atoms, particles or orbits are involved.\\
{\bf Software}	\\
- Concepts which are in essence software (e.g., video games, matchboxes) belong to 
the Software domain. \\
{\bf Sport}	\\
- The biography of a sportsman should be considered within the Sport domain. \\
- An article of a location with a section on Sport does not belong to the domain 
Sport.
\end{minipage}
}
\medskip

We paid 0.06~USD per HIT, 
each of which consisted of 10 binary annotations, and 
set a minimum working time of 120 seconds. We manually annotated $10\%$ of the 
instances for quality control and requested an annotation accuracy of $80\%$ to 
verify the annotation quality. Each item was judged three times.

\end{document}